%% file: main.tex
\documentclass[10pt,twocolumn,letterpaper]{article}

\usepackage[pagenumbers]{iccv} 


\definecolor{iccvblue}{rgb}{0.21,0.49,0.74}
\usepackage[pagebackref,breaklinks,colorlinks,allcolors=iccvblue]{hyperref}
\usepackage{multirow}
\usepackage{comment}
\usepackage{colortbl} 
\usepackage{pgffor}
\usepackage{wrapfig}

\definecolor{tabfirst}{rgb}{1, 0.7, 0.7} 
\definecolor{tabsecond}{rgb}{1, 0.85, 0.7} 
\definecolor{tabthird}{rgb}{1, 1, 0.7} 

\def\paperID{10405} 
\def\confName{ICCV}
\def\confYear{2025}

\title{Sparfels: Fast Reconstruction from Sparse Unposed Imagery}

\author{
Shubhendu Jena\textsuperscript{*},
Amine Ouasfi\textsuperscript{*},
Mae Younes,
Adnane Boukhayma\\
Inria, Univ. Rennes, CNRS, IRISA
}

\begin{document}
\twocolumn[{%
\renewcommand\twocolumn[1][]{#1}%
\vspace{-4em}
\maketitle
\vspace{-2.5em}
\includegraphics[width=1.0\linewidth]{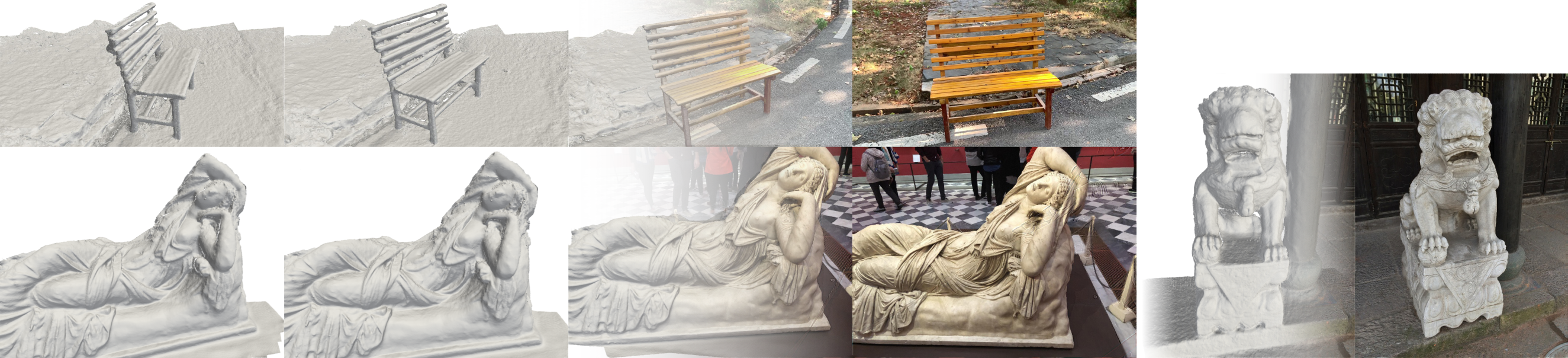}
\vspace{-2em}
\captionof{figure}{Examples of reconstruction meshes  obtained within {\bf 3 minutes} from sparse pose-free images via our method. We used here {\bf6 (left) and 3 (right) input images} respectively from scenes of datasets MVImgNet \cite{yu2023mvimgnet} and BlendedMVS \cite{yao2020blendedmvs}.  
\vspace{1em}}
\label{fig:tease}
}]

\renewcommand*{\thefootnote}{\fnsymbol{footnote}}
\footnotetext[1]{Equal contribution.}


\input{sec/abstract}  
\vspace{-7pt}
\input{sec/intro}
\vspace{-5pt}
\input{sec/relatedwork}

\vspace{-5pt}
\input{sec/method}
\vspace{-5pt}
\input{sec/results}

\vspace{-5pt}
\input{sec/conc}

\clearpage
\appendix
\renewcommand{\thesection}{\Alph{section}}

{
    \small
    \bibliographystyle{ieeenat_fullname}
    \bibliography{main}
}

\clearpage

\twocolumn[{
\vspace{2ex}
\begin{center}
    {\Large\bfseries Sparfels: Fast Reconstruction from Sparse Unposed Imagery}\\[0.5ex]
    {\large\bfseries – Supplementary Material –}\\[1.5ex]
    {\normalsize
    Shubhendu Jena\textsuperscript{*},
    Amine Ouasfi\textsuperscript{*},
    Mae Younes,
    Adnane Boukhayma\\
    Inria, Univ. Rennes, CNRS, IRISA\\[1ex]
    }
\end{center}
\vspace{2ex}
}]
\renewcommand*{\thefootnote}{\fnsymbol{footnote}}
\footnotetext[1]{Equal contribution.}

\input{sec/supp}

\end{document}

%% file: sec/abstract.tex
\begin{abstract}
We present a method for \textbf{Spar}se view reconstruction with sur\textbf{f}ace \textbf{el}ement splatting that runs within 3 minutes on a consumer grade GPU.  While few methods address sparse radiance field learning from noisy or unposed sparse cameras, shape recovery remains relatively underexplored in this setting. Several radiance and shape learning test-time optimization methods address the sparse posed setting by learning data priors or using combinations of external monocular geometry priors. Differently, we propose an efficient and simple pipeline harnessing a single recent 3D foundation model. We leverage its various task heads, notably point maps and camera initializations to instantiate a bundle adjusting 2D Gaussian Splatting (2DGS) model, and image correspondences to guide camera optimization midst 2DGS training. Key to our contribution is a novel formulation of splatted color variance along rays, which can be computed efficiently. Reducing this moment in training leads to more accurate shape reconstructions. We demonstrate state-of-the-art performances in the sparse uncalibrated setting in reconstruction and novel view benchmarks based on established multi-view datasets. Code will be made available at \href{https://shubhendu-jena.github.io/Sparfels-web/}{https://shubhendu-jena.github.io/Sparfels-web/} 
\end{abstract}

%% file: sec/intro.tex
\section{Introduction}
\label{sec:intro}

3D reconstruction is a long standing problem at the intersection of computer vision, graphics and machine learning. It is core to a myriad of downstream applications in virtual and augmented reality, autonomous driving, medical imaging, film and animation, to name a few. Classical optimization based techniques such as structure from motion and multi-view stereo \cite{sfm,mvs}, shape from visual cues such as shading or silhouette \cite{sfshading,sfsilhouette}, photometric stereo \cite{PS} \etc, have been popular for decades in this regard. More recently, deep learning has been progressively disrupting 3D vision and top-down approaches to vision more broadly. Among the latest additions, large vision and foundation models (\eg DUSt3R \cite{wang2024dust3r}, MASt3R \cite{leroy2024grounding}) can offer robustness and fast feed-forward inference compared to classical photogrammetry \cite{sfm,mvs}.
Given a few raw uncalibrated images of a scene, they are capable of inferring reasonable relative poses and a competitive coarse 3D structure. These predictions can lack precision and details, though, and cannot enable stand-alone novel view synthesis. 


Furthermore, state-of-the-art data-prior free scene specific learning methods for reconstruction can deliver highly detailed geometry and impressive novel view synthesis capabilities from dense calibrated image grids. While implicit and some grid shape representations (\eg NeuS, VolSDF, HF-NeuS, Voxurf \cite{wang2021neus,yariv2021volume,wang2022hf,wu2023voxurf}) trained via differentiable volumetric rendering (NeRF \cite{mildenhall2021nerf}) have been powerful tools for this task for a while, Gaussian Splatting \cite{kerbl20233d} based alternatives emerged with comparable novel view and reconstruction \cite{huang20242d,yu2024gaussian} quality, within a fraction of the training time, and with orders of magnitude higher interactive rendering frame rates. These methods typically fail however under uncalibrated and sparse training images. Contemporary literature attempts to hedge against the underconstrained nature of the sparse view problem through regularization with a multitude of deep geometric priors (monocular depth \cite{deng2022depth,wang2023sparsenerf,zhu2023fsgs}, normal \cite{yu2022monosdf}, point cloud \cite{raj2024spurfies} \etc), or training new priors from scratch \cite{long2022sparseneus,raj2024spurfies,xu2024sparp}. Such strategies can prove to be cumbersome, impractical or costly in terms of data, memory or compute.

In this paper, we propose to marry the best of both worlds efficiently — namely pre-exisiting large 3D vision feed-forward models (MASt3R) and fast test-time optimization (2DGS) — with the aim of {\bf addressing  the underexplored problem of 3D geometric reconstruction from sparse unposed images within minutes} (less than 3 on average in our experiments) at minimal cost.  

We propose to exploit a single pre-existing foundation model (MASt3R \cite{leroy2024grounding}) to the fullest to bootstrap our reconstruction, without needing to load or deploy any additional deep neural networks. Given a few sparse color images, MASt3R delivers a coarse point cloud and initial camera poses. We use these to initialize a bundle adjusting 2D Gaussian Splatting (2DGS) optimization via differentiable render-and-compare, where surface elements are instantiated at the point samples, oriented initially according to their normals. Differently from the closest competition \cite{fan2024instantsplat1, fan2024instantsplat2}, we harness the MASt3R \cite{leroy2024grounding} produced image pair correspondences to guide camera optimization during 2DGS training. This is achieved via a splatted depth re-projection error loss. The resulting camera alignment improvement proves crucial for obtaining accurate shape representations. 

By analyzing the surface element splatting from a statistical moment perspective, rendered pixels can be seen as color expectation over fragments following the ray termination probability. Based on this, we derive an expression of splatted color variance along the ray, and propose a novel loss tasked with reducing it. This results in improved reconstruction metrics and qualitative sharper mesh details as witnessed by our ablation studies. 

The integration of these elements yields state-of-art performance in sparse pose-free 3D reconstruction, obtained in very short times on mid-range GPUs. We evaluate our method using established reconstruction and novel view datasets \cite{aanaes2016large,yao2020blendedmvs,knapitsch2017tanks,yu2023mvimgnet,barron2022mip}, showcasing both superior geometric reconstruction, camera pose estimation and novel view synthesis \wrt comparable state-of-the-art baselines.

Evaluation of 3D shape reconstruction in the posed setting relies traditionally on mesh Chamfer distance to ground-truth, as used in multi-view reconstruction benchmarks like DTU \cite{aanaes2016large}. This is a mesh pose and scale sensitive metric. In the unposed setting, reconstructions are obtained naturally up-to a rigid transformation. RANSAC powered alignments between predicted and ground-truth cameras have been shown to be acceptable for pre-evaluation reconstruction mesh alignment in the dense view setting \cite{bian2023porf}, and good enough for camera pose evaluation for sparse sets \cite{truong2023sparf}. However, we found these camera based alignments to be too inaccurate in the sparse setting to enable reliable mesh pre-alignments for Chamfer distance evaluation. Hence, inspired by the rigid-transform invariant evaluation in \cite{wang2024dust3r}, we compare the reconstructed mesh depth and normal rasterizations to the groundtruth in screen space in our evaluation, and show considerable improvements compared to the competition in the standard DTU benchmark (Tab.\ref{tab:rel},\ref{tab:nc},\ref{tab:ate}). Further qualitative comparison in BlendedMVS \cite{yao2020blendedmvs}, MVImgNet \cite{yu2023mvimgnet}, Mip-NeRF 360 \cite{Barron2022MipNeRF3U} illustrate the quality of our reconstructions.




\begin{figure}[t!]
    \centering
    \includegraphics[width=1.0\linewidth]{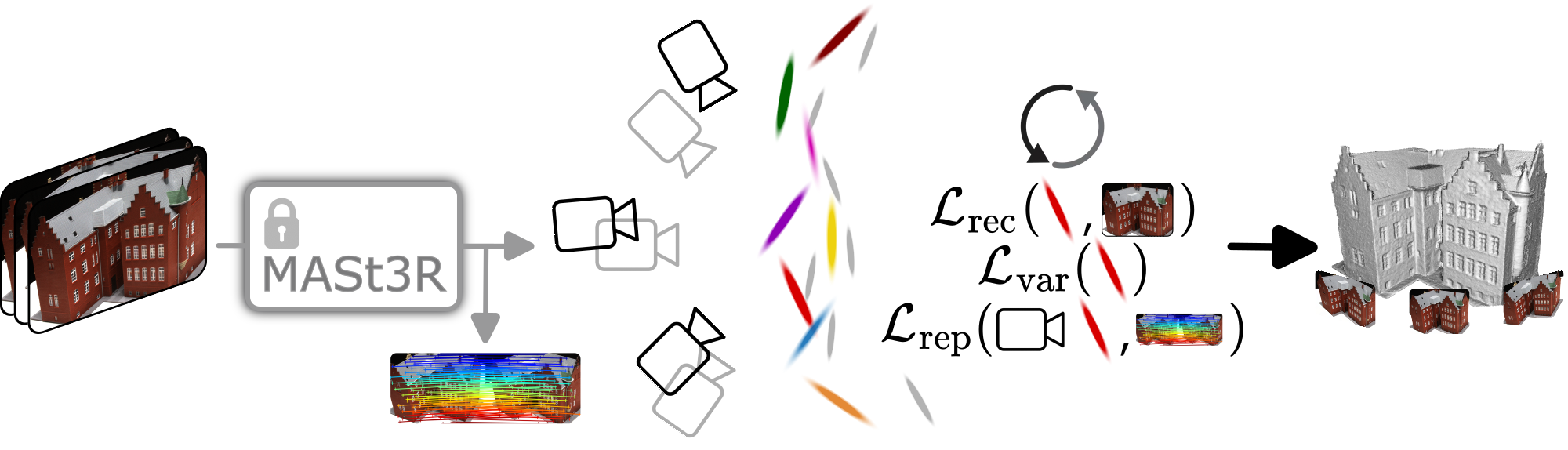} 
    \caption{3D Foundation models meet differentiable perspective accurate Splatting for fast reconstruction from pose-free sparse images. We bootstrap a bundle adjusting 2DGS with MASt3R \cite{leroy2024grounding} 3D point clouds and initial camera predictions. Besides standard 2DGS photometric reconstruction and geometric regularization losses, we leverage a camera reprojection error loss using MASt3R's correspondences and a novel splatted color variance reduction loss.}
    \vspace{-10pt}
    \label{fig:abl}
\end{figure}

%% file: sec/relatedwork.tex
\section{Related work} 
\label{sec:related}

\paragraph{NVS and 3D Reconstruction} Novel view synthesis (NVS) aims to generate unseen views from input images~\cite{avidan1997novel,mildenhall2019local}. Neural Radiance Fields (NeRF)~\cite{mildenhall2021nerf} revolutionized this field by representing 3D scenes as continuous implicit functions, though challenges remain in efficiency and quality~\cite{Barron2022MipNeRF3U,barron2022mip,barron2023zip,penner2017soft,seitz2002photorealistic,srinivasan2020lighthouse}. Inspired by the success of other explicit representations 
(\eg meshes \cite{wang2018pixel2mesh,kato2018neural,jena2022neural} and point clouds \cite{fan2017point,aliev2020neural}), recent advancements introduced primitive-based scene representations~\cite{chen2023neurbf,kerbl20233d,xu2022point}, notably 3D Gaussian Splatting (3DGS)~\cite{kerbl20233d}, which uses anisotropic 3D Gaussians with differentiable splatting for efficient high-quality reconstruction. This has sparked extensions for surface reconstruction~\cite{huang20242d,yu2024gaussian,zhang2024rade}, neural representations~\cite{lu2024scaffold}, multi-resolution modeling~\cite{yu2024mip,feng2024flashgs}, and feed-forward methods~\cite{charatan2024pixelsplat,chen2025mvsplat,fan2024large,xu2024depthsplat,hong2024pf3plat}.
In parallel, implicit representations model 3D geometry as continuous functions~\cite{darmon2022improving,jiang2020sdfdiff,kellnhofer2021neural,niemeyer2020differentiable,oechsle2021unisurf}. Methods like IDR~\cite{yariv2020multiview}, NeuS~\cite{wang2021neus}, and VolSDF~\cite{yariv2021volume} enhance geometric accuracy but often suffer from slow rendering. In contrast, 2D Gaussian Splatting (2DGS)~\cite{huang20242d} offers real-time rendering while improving view consistency for surface reconstruction through planar Gaussian primitives.

\paragraph{Sparse-View NVS and 3D Reconstruction} Traditional NeRFs and GS methods require carefully captured images and SfM preprocessing with COLMAP~\cite{schonberger2016structure}, which could limit practical applications. The few-shot setup has been addressed for both reconstruction from Point Cloud \cite{dro, nap, sparseocc, ntps} and images. 
Various approaches address the challenge of limited views through regularization techniques, including depth supervision~\cite{deng2021depth,niemeyer2021regnerf,wang2023sparsenerf,rgbd}, foundation model supervision~\cite{jain2021putting,xu2022sinnerf,wu2023reconfusion}, specialized architectures~\cite{yu2021pixelnerf,yang2023freenerf}, while SparseCraft~\cite{younes2024sparsecraft} uses learning-free MVS cues for SDF regularization. For Gaussian Splatting, methods like FSGS~\cite{zhu2023fsgs} and SparseGS~\cite{xiong2023sparsegs} incorporate external priors. Generalizable methods address sparse-view challenges by pretraining on large datasets, enabling feed-forward inference. Inputs include images (\eg ~\cite{charatan2024pixelsplat,chen2025mvsplat,liu2024fast,genlf,sparsplat}) and 3D Points (\eg ~\cite{boulch2022poco,pcp,fssdf,nksr,robust,mixing}).
For surface reconstruction from color images, approaches like SparseNeuS~\cite{long2022sparseneus}, VolRecon~\cite{ren2023volrecon}, ReTR~\cite{liang2024retr}, UfoRecon~\cite{na2024uforecon}, GeoTransfer~\cite{jena2024geotransfer}, and Spurfies~\cite{raj2024spurfies} leverage various mechanisms such as using image priors to form multi-view cost volumes, using transformer architectures to enhance reconstruction from limited views, as well as using reconstruction priors trained on sparse point clouds.

\paragraph{Pose-Free NVS and 3D Reconstruction} Pose-free methods eliminate the need for accurate camera calibration. NeRFmm~\cite{wang2021nerf}, BARF~\cite{lin2021barf}, SCNeRF~\cite{jeong2021self}, and GARF~\cite{chng2022gaussian} jointly optimize camera parameters and scene representations. SPARF~\cite{truong2023sparf} and PoRF~\cite{bian2023porf} address optimization with noisy poses, though often requiring lengthy training. Recent work like Nope-NeRF~\cite{bian2023nope}, Lu-NeRF~\cite{cheng2023lu}, LocalRF~\cite{meuleman2023progressively}, and CF-3DGS~\cite{fu2023colmap} leverage depth information but typically assume dense video sequences. InstantSplat~\cite{fan2024instantsplat1,fan2024instantsplat2} addresses these limitations by using MASt3R~\cite{leroy2024grounding} for initialization and jointly optimizing camera poses and 3D models.
Our work builds on these foundations but focuses specifically on using 2D Gaussian Splatting for efficient 3D surface reconstruction from sparse, uncalibrated views. We initialize with MASt3R~\cite{leroy2024grounding} and refine through joint optimization of Gaussian parameters and camera poses, incorporating a novel variance-based color loss for improved detail preservation. Unlike previous approaches requiring dense sequences, known intrinsics, or long optimization times, our method efficiently handles sparse, uncalibrated images while achieving high-quality reconstructions.

%% file: sec/method.tex
\section{Method}

Given a few unposed color images $\{I^i\}_{i=1}^N$, our goal is to recover a 3D triangle mesh $\mathcal{S}$ and render novel views of the observed scene efficiently. We achieve this by fitting a bundle adjusting 2DGS \cite{huang20242d} model to the images of the scene, bootstrapped by point map and correspondence feedforward predictions from a 3D foundation model (MASt3R \cite{leroy2024grounding}) and carefully designed regularizations. At test time, mesh $\mathcal{S}$ can be extracted using the TSDF algorithm in Open3D \cite{zhou2018open3d} using the converged 2DGS model rendered depth, as recommended in \cite{huang20242d}. We detail the workings of our method below.





\subsection{Preliminaries}
\paragraph{2D Gaussian Splatting}

Our method builds upon 2D Gaussian Splatting (2DGS)~\cite{huang20242d}, which represents 3D scenes using planar disk primitives defined within local tangent planes. Unlike volumetric 3D Gaussians~\cite{kerbl20233d}, 2DGS is a surface-based representation that improves multi-view consistency. Each 2D Gaussian is defined as:

\begin{equation} 
G(\mathbf{p}) = \exp\left(-\frac{u(\mathbf{p})^2 + v(\mathbf{p})^2}{2}\right) 
\end{equation}

where $u(\mathbf{p})$ and $v(\mathbf{p})$ are local UV space coordinates of a pixel $\mathbf{p}$ in the primitive's tangent plane. This ray-splat intersection can be expressed as a function of the camera parameters and splat attributes differentiably. Each primitive is parameterized by the center of the Gaussian in world coordinates $\mathbf{x} \in \mathbb{R}^3$, a quaternion representing the orientation of the primitive $\mathbf{q} \in \mathbb{R}^4$, scale factors for the two dimensions in the local UV tangent plane $\mathbf{s} \in \mathbb{R}^2$, opacity $\alpha \in \mathbb{R}$ controlling the transparency of the primitive, and Spherical harmonic coefficients ${\mathbf{c_{\text{sh}}}_k \in \mathbb{R}^3 | k=1,\hdots,n}$ encoding view-dependent appearance. Rotation $\mathbf{q}$ defines a local coordinate frame $[\mathbf{u}, \mathbf{v}, \mathbf{n}]$, $\mathbf{n}$ being the normal vector orthogonal to the primitive's plane, and $\mathbf{u}$ and $\mathbf{v}$ spanning the tangent plane.

For rendering, 2DGS uses alpha-compositing for volumetric rendering similarly to 3DGS. Given a set of primitives along a ray sorted from front to back ${G_1, G_2, ..., G_K}$, the final color $C$ is computed as:
\begin{equation} C(\mathbf{p}) = \sum_{i=1}^K \alpha'_i \mathbf{c}_i \prod_{j=1}^{i-1}(1-\alpha'_j), \end{equation}
where $\alpha'_i=\alpha_iG(\mathbf{p})$ is the modified opacity of the $i$-th primitive and $\mathbf{c}_i$ is its color, computed from the spherical harmonic coefficients based on the viewing direction.






\paragraph{MASt3R}

MASt3R~\cite{leroy2024grounding} is a foundation model for 3D vision that extends DUSt3R~\cite{wang2024dust3r} with more powerful feature extraction and matching capabilities. It tackles the multi-view stereo reconstruction task without requiring prior camera calibration or poses.

Given a pair of images $I^1, I^2 \in \mathbb{R}^{W \times H \times 3}$, MASt3R outputs two corresponding pointmaps $\mathbf{P}_{1,1}, \mathbf{P}_{2,1} \in \mathbb{R}^{W \times H \times 3}$ with associated confidence maps $\mathbf{O}_{1,1}, \mathbf{O}_{2,1} \in \mathbb{R}^{W \times H}$. Each pointmap assigns a 3D point to every pixel in the image. Importantly, both pointmaps are expressed in the same coordinate frame of the first image $I^1$.

The architecture of MASt3R consists of a Siamese ViT encoder that processes both images independently, Transformer decoders that exchange information via cross-attention, and regression heads that output pointmaps and confidence maps. MASt3R is trained using a confidence-weighted regression loss:
\begin{equation} 
\mathcal{L}_{\text{conf}} = \sum_{v \in {1,2}} \sum_{i} \mathbf{O}_{v,1}^i \mathcal{L}_{\text{reg}}(v,i) - \alpha \log \mathbf{O}_{v,1}^i,
\end{equation}
where $\mathcal{L}_{\text{reg}}(v,i)$ measures the Euclidean distance between predicted and ground-truth 3D points for pixel $i$, and $\alpha$ is a regularization parameter.

Besides pointmaps, MASt3R also provides dense pixel-wise correspondences between the input views. These correspondences establish reliable geometric constraints across different viewpoints, which we leverage in our optimization framework.

\subsection{Scene Initialization}
\paragraph{Global Geometry Alignment}

Given a set of input images ${I^1, I^2, ..., I^N}$, we first construct a connectivity graph $\mathcal{G}(\mathcal{V}, \mathcal{E})$ where each vertex $v \in \mathcal{V}$ represents an image, and each edge $e = (n, m) \in \mathcal{E}$ indicates a pair of images with shared visual content. For each edge in the graph, we run MASt3R to obtain pairwise pointmaps and correspondences. To align these pairwise predictions into a globally consistent reconstruction, we follow the optimization approach from MASt3R:
\begin{equation} 
\min_{\chi, P, \sigma} \sum_{e \in \mathcal{E}} \sum_{v \in e} \sum_{i=1}^{HW} \mathbf{O}_{v,e}^i |\chi_i^v - \sigma_e P_e \mathbf{P}_{v,e}^i|, 
\end{equation}
where $\chi_i^v$ is the globally aligned pointmap for view $v$ at pixel $i$, $P_e \in \mathbb{R}^{3 \times 4}$ the rigid transformation for edge $e$, $\sigma_e > 0$ is a scaling factor for edge $e$, $\mathbf{O}_{v,e}^i$ the confidence value for pixel $i$ in view $v$ for edge $e$, and finally $\mathbf{P}_{v,e}^i$ the original pointmap from MASt3R.
This optimization produces globally aligned pointmaps $\{{\chi^v}^*\}_{v=1}^N \in \mathbb{R}^{H \times W \times 3}$ and camera parameters (intrinsics and extrinsics for each view $v$). 
We stabilize the estimated focal lengths by averaging them across all training views: $\bar{f} = \frac{1}{N} \sum_{v=1}^N f_i^*$, assuming a single-camera setup.



\vspace{-10pt}
\paragraph{Primitive Initialization}

We leverage the merged globally aligned pointmap to initialize 2D Gaussian primitives with suitable orientation, which is crucial for accurately representing surface geometry. For each point $\mathbf{p}_i$ in the globally aligned pointmap, we compute a local surface normal using Principal Component Analysis (PCA). First, we determine the $k$-nearest neighbors $\{\mathbf{p}_j\}_{j=1}^k$ and compute the covariance matrix $\mathbf{C} = \frac{1}{k}\sum_{j=1}^{k} (\mathbf{p}_j - \bar{\mathbf{p}})(\mathbf{p}_j - \bar{\mathbf{p}})^T$, where $\bar{\mathbf{p}}$ is the centroid of the neighborhood. We then perform eigendecomposition on $\mathbf{C}$ to obtain the eigenvectors $\{\mathbf{v}_1, \mathbf{v}_2, \mathbf{v}_3\}$ and corresponding eigenvalues $\{\lambda_1, \lambda_2, \lambda_3\}$, sorted such that $\lambda_1 \geq \lambda_2 \geq \lambda_3$. The eigenvector $\mathbf{v}_3$, associated with the smallest eigenvalue, is chosen as the normal direction, i.e., $\mathbf{n} = \mathbf{v}_3$. To construct an orthonormal basis, we initialize an arbitrary vector $\mathbf{a}$ orthogonal to $\mathbf{n} = (n_x, n_y, n_z)$, defined as $\mathbf{a} = (-n_y, n_x, 0)^T$. The tangent vector is then computed as $\mathbf{u} = \frac{\mathbf{n} \times \mathbf{a}}{\|\mathbf{n} \times \mathbf{a}\|}$ and the bitangent as $\mathbf{v} = \mathbf{n} \times \mathbf{u}$. Finally, the local reference frame $[\mathbf{u}, \mathbf{v}, \mathbf{n}]$ is used to define a rotation matrix, which can be converted to a quaternion $\mathbf{q}_i$ for efficient representation. Next, we initialize each 2D Gaussian primitive with position $\mathbf{x}_i$ directly from the pointmap, rotation $\mathbf{q}_i$, scale $\mathbf{s}_i$ set proportionally to the local pointmap density, and opacity $\alpha_i$ set to a default value (e.g. 0.8). Color for the lowest spherical harmonic band (DC component) is initialized from the original image colors projected onto the point, with higher-order coefficients set to zero. This initialization preserves the surface structure from the MASt3R reconstruction while enabling efficient optimization with 2D Gaussian primitives.

\subsection{Joint Optimization}
\paragraph{Correspondence Loss}
\label{ref:corr_section}
Using the dense correspondences provided by MASt3R, and inspired by SPARF~\cite{truong2023sparf}, we formulate a multi-view correspondence loss that enforces geometric consistency across different viewpoints:
\begin{equation} \mathcal{L}_{\text{corr}} =  w_{p_n,p_m} \rho\left(p_m - \pi\left(P_m^{-1} P_n \pi^{-1}(p_n, d_n)\right)\right), \end{equation}

where $(p_n, p_m)$ is a correspondence pair with $p_n \in [0,1]^2$ a pixel in view $n$ and $p_m$ its corresponding pixel in view $m$. $w_{p_n,p_m} \in [0,1]$ is the confidence of the correspondence. $\rho$ represents the Huber loss \cite{huber1992robust} function. $\pi: \mathbb{R}^3 \rightarrow \mathbb{R}^2$ is the projection operator mapping 3D points to pixel coordinates.  $\pi^{-1}: \mathbb{R}^2 \times \mathbb{R} \rightarrow \mathbb{R}^3$ is the back-projection operator mapping pixel coordinates and depth to 3D points. $d_n$ is the 2DGS model splatted depth at pixel $p_n$. $P_n$ and $P_m$ represent the camera parameters for views $n$ and $m$. This loss enforces that corresponding pixels across views should project to the same 3D point when back-projected using their respective depths and camera parameters, and thus helps guide our bundle adjustment to a good minima.
\vspace{-10pt}
\paragraph{Photometric Loss}

Following 3DGS and 2DGS, we use a combination of $\mathcal{L}_1$ image reconstruction loss and structural similarity (SSIM) to optimize the rendering of our system: 
\begin{equation}
\mathcal{L}_{\text{photo}} = (1-\lambda) \cdot \mathcal{L}_1 + \lambda \cdot \mathcal{L}_{\text{SSIM}} + \mathcal{L}_{\text{reg}}.
\end{equation}
Notice that 2DGS comes also with geometric regularization term $\mathcal{L}_{\text{reg}}$ binding pixel depth normals to splatted normals, and reducing depth distortion, as detailed in \cite{huang20242d}.




\vspace{-10pt}
\paragraph{Variance Regularization Loss}
\label{ref:var_reg_section}
A key contribution of our method is the introduction of a splatted color variance regularization loss. The fundamental insight is that volume rendering can be interpreted as computing the expected value of colors sampled along a ray \cite{tagliben}:
\begin{equation}
C = \mathbb{E}_{\bold t\sim p(\bold t)}[c( \bold t)] = \sum_{i=1}^K \alpha_i \mathbf{c}_i \prod_{j=1}^{i-1}(1-\alpha_j),
\end{equation}
where $c(\bold t)$  represents the color of a point $\bold t$ along the ray, and the expectation is taken with respect to $p(t) = \sigma( t) T(t)$ with  $\sigma(t)$ being the density at $ t$ and $T(t)$ representing the transmittance function. 

We hypothesize that stabilizing and robustifying our training would lead to more robust and improved reconstructions. In order to robustify the training, we would like to control how the rendered color deviates from the ground truth color when the geometry is perturbed, \ie given divergence measure $\mathcal{D}$, we want to minimize the following worst-case loss along the ray:
\begin{equation} 
\label{eqn:var_loss1}
\mathcal{L}_r= \sup_{q:D(q,p) < \rho}|{\mathbb{E}_{\bold t\sim q(\bold t)}[c( \bold t)] - C_{gt}}|,
\end{equation}
where $\rho$ controls the strength of the perturbation. By leveraging theorem 2 in \cite{vardro}, this loss can be upper bounded by:
\begin{equation} 
\label{eqn:var_loss2}
  \mathcal{L}_1 + \eta \sqrt{\mathbb{V}\text{ar}_{\boldsymbol{t} \sim p(\boldsymbol{t})}[c(\boldsymbol{t})]}.
\end{equation}
Consequently, we consider minimizing the variance of the color along the ray under the distribution $p(t)$ that can be expressed as follows:
\begin{align} 
\label{eqn:var_loss3}
 \mathcal{L}_\text{var} &=
 \mathbb{V}ar_{\bold t\sim p(\bold t)}[c( \bold t)]\\ 
 &= \mathbb{E}_{\bold t\sim p(\bold t)}[ (c( \bold t)-C)^2] = \mathbb{E}_{\bold t\sim p(\bold t)}[c( \bold t)^2] - C^2. 
\end{align}
%
For efficient implementation, we compute the first term  using a modified CUDA kernel of the 2DGS splatting algorithm, where we render the square of colors in addition to the rendering of the colors. This allows for efficient forward and backward passes during optimization. Intuitively,  reducing rendered color variance reduces uncertainty in the rendering, which leads to sharper and more multi-view coherent reconstructions. This reflects positively on the qualitative (Fig \ref{fig:abl}) and quantitative (Tab \ref{tab:abl_loss} and Fig \ref{fig:abl_no_views}) performance of our reconstructions. We note that in practice, loss $\mathcal{L}_\text{var}$ is averaged across pixels for a given view.  

\vspace{-10pt}
\paragraph{Training Objective}

Our total optimization objective combines the three loss terms:
\begin{equation} 
\mathcal{L} = \lambda_{\text{photo}}\mathcal{L}_{\text{photo}} + \lambda_{\text{corr}}\mathcal{L}_{\text{corr}} + \lambda_{\text{var}}\mathcal{L}_{\text{var}}, \end{equation}
where $\lambda_{\text{photo}}$, $\lambda_{\text{corr}}$, and $\lambda_{\text{var}}$ are weighting factors that balance the contribution of each term. We optimize all parameters simultaneously using the Adam \cite{kingma2014adam} optimizer with a learning rate scheduling following \cite{huang20242d}. Unlike multi-stage approaches \cite{truong2023sparf}, our single-stage optimization allows for faster convergence while maintaining high reconstruction quality. The optimization updates both camera extrinsics $\{P\}$ per input view and 2D Gaussian primitive parameters $\{\alpha, \mathbf{c}, \mathbf{q}, \mathbf{s}\}$ according to the gradient of the total loss.

%% file: sec/results.tex
\section{Experiments} \label{sec:experiments}


We evaluate our approach across 3D reconstruction, novel view synthesis, and camera pose estimation. We outline the experimental setup, including datasets, baseline methods, and implementation details, followed by quantitative and qualitative comparisons. Ablation studies further analyze the impact of key components.  

\subsection{Datasets and Experimental Configuration}  
Our method is tested on benchmark datasets~\cite{aanaes2016large,yao2020blendedmvs,knapitsch2017tanks,yu2023mvimgnet,barron2022mip}. Further implementation details and additional qualitative results are provided in the supplementary material.  

\noindent\textbf{Few-Shot 3D Reconstruction:}  
We evaluate sparse 3D reconstruction on \textbf{DTU}~\cite{aanaes2016large}, \textbf{BlendedMVS}~\cite{yao2020blendedmvs}, \textbf{MVImgNet}~\cite{yu2023mvimgnet}, and \textbf{MipNeRF360}~\cite{barron2022mip}. For 3-view reconstruction on \textbf{DTU}, we follow SparseNeuS~\cite{long2022sparseneus}, evaluating 15 scenes with two sets of 3-view inputs, reporting the mean per scene. We also provide qualitative comparisons for 3, 6, and 12-view settings on selected scenes from \textbf{BlendedMVS}~\cite{yao2020blendedmvs}, \textbf{MVImgNet}~\cite{yu2023mvimgnet}, and \textbf{MipNeRF360}~\cite{barron2022mip}.  

\noindent\textbf{Few-Shot Novel View Synthesis:}  
We conduct novel view synthesis (NVS) on \textbf{Tanks and Temples}~\cite{knapitsch2017tanks} (8 scenes), \textbf{MVImgNet}~\cite{yu2023mvimgnet} (7 scenes), and \textbf{MipNeRF360}~\cite{barron2022mip} (9 scenes), following InstantSplat~\cite{fan2024instantsplat1, fan2024instantsplat2}. Results are reported under sparse input settings (3, 6, and 12 views).  

\noindent\textbf{Camera Pose Estimation:}  
Camera poses are evaluated on \textbf{Tanks and Temples}~\cite{knapitsch2017tanks}, \textbf{MVImgNet}~\cite{yu2023mvimgnet}, and \textbf{MipNeRF360}~\cite{barron2022mip}, following InstantSplat~\cite{fan2024instantsplat1, fan2024instantsplat2}. Additionally, for few-shot reconstruction, we assess camera poses on \textbf{DTU}~\cite{aanaes2016large} using the same 3-view evaluation protocol as in the reconstruction task.



\vspace{-10pt}
\paragraph{\textbf{Evaluation Metrics.}} For 3D reconstruction quality, we employ the Absolute Relative Error (Rel) and Normal Consistency (NC). Rel measures the accuracy of depth estimation following DUSt3R~\cite{wang2024dust3r}, while NC evaluates the geometric fidelity by measuring the alignment between predicted and ground truth surface normals. We apply these metrics to our final meshes. We compute ground truth normals from the gradients of the ground truth DTU depth maps.

For novel view synthesis, we use standard metrics: Peak Signal-to-Noise Ratio (PSNR), Structural Similarity Index Measure (SSIM) \cite{wang2004image}, and Learned Perceptual Image Patch Similarity (LPIPS) \cite{zhang2018unreasonable}. We note that for this evaluation we follow InstantSplat\cite{fan2024instantsplat1, fan2024instantsplat2}: for novel test views without pre-computed camera poses, we perform an additional camera optimization step, which optimizes the test cameras while keeping the 2D Gaussian primitives fixed using the $\mathcal{L}_1$ photometric reconstruction loss.

For camera pose estimation, we report the Absolute Trajectory Error (ATE) as defined in~\cite{fan2024instantsplat1,bian2023nope}, using COLMAP-computed poses from dense views as ground-truth references.
\vspace{-10pt}
\paragraph{\textbf{Baselines.}} We compare our method against several state-of-the-art approaches:

\begin{itemize} 
\item \textbf{NVS specific methods}: Pose-free methods NoPe-NeRF~\cite{bian2023nope}, CF-3DGS~\cite{fu2023colmap}, NeRFmm~\cite{wang2021nerf}, SPARF \cite{truong2023sparf} and InstantSplat~\cite{fan2024instantsplat1,fan2024instantsplat2} (in both S and XL configurations). We also include comparisons with 3DGS~\cite{kerbl20233d} and FSGS~\cite{zhu2023fsgs}, which use pre-computed camera parameters from Colmap \cite{schonberger2016structure}.  
\item \textbf{Reconstruction-specific methods}: MASt3R~\cite{leroy2024grounding}, Colmap \cite{schonberger2016structure}, UfoRecon~\cite{na2024uforecon}, CasMVSNet~\cite{gu2020cascade}, Spurfies~\cite{raj2024spurfies}, SpaRP \cite{xu2024sparp}, SparseCraft~\cite{younes2024sparsecraft}, InstantSplatGOF (our implementation based on \cite{fan2024instantsplat2}) and InstantSplat2dgs \cite{fan2024instantsplat2}. Of these, similar to our approach, InstantSplatGOF (our implementation based on \cite{fan2024instantsplat2}) and InstantSplat2dgs \cite{fan2024instantsplat2} further refine cameras. 
\end{itemize}

\vspace{-10pt}
\paragraph{\textbf{Implementation Details.}} Our framework is implemented in PyTorch. We set the optimization iterations to 1k for DTU dataset reconstruction experiments and between 2k-4k for other reconstruction and novel view synthesis experiments. For test-time optimization, camera parameters are jointly optimized for 1k iterations (while keeping the learned Gaussian parameters fixed). We initialize our pipeline with MASt3R~\cite{leroy2024grounding} configured at a resolution of 512 pixels and use Open3D~\cite{zhou2018open3d} to compute normals for the point cloud, which are used to initialize rotation parameters of the 2D Gaussians. All experiments are conducted on a single NVIDIA A6000 GPU. We set $\lambda_{\text{photo}} = 1.0$ and $\lambda_{\text{corr}} = 5 \times 10^{-5}$. The variance loss weight $\lambda_{\text{var}}$ follows a cosine annealing schedule: 
\(\lambda_{\text{var}} = 0.5 \left(1 + \cos \left( \frac{\pi \cdot t}{T} \right) \right)\), where $t$ denotes the current training step and $T$ the total number of steps.
 This schedule gradually decreases $\lambda_{\text{var}}$ from 1.0 to 0.0, ensuring an adaptive regularization throughout training.    

\subsection{3D Reconstruction Evaluation} \label{subsec:reconstruction}
For few-shot reconstruction, we follow the evaluation protocol from prior work~\cite{long2022sparseneus,younes2024sparsecraft,na2024uforecon}, cleaning the generated meshes with masks from training views. We evaluate the Absolute Relative Error (Rel) and Normal Consistency (NC) between predicted depths/normals from the extracted meshes and the ground truth depths/normals from the DTU dataset~\cite{aanaes2016large}.

\begin{figure}[t] \centering \includegraphics[width=\linewidth]{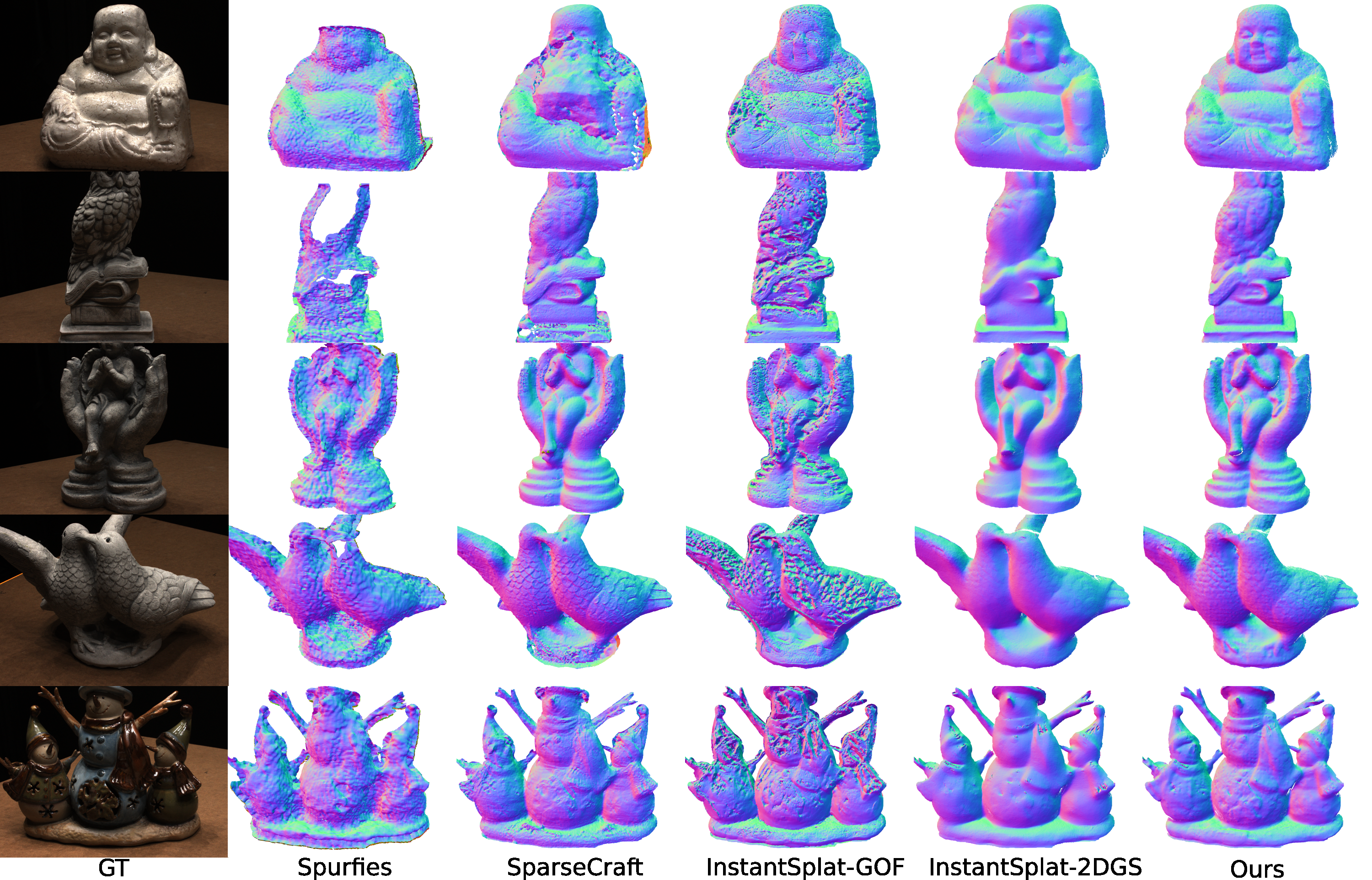} 
\vspace{-15pt}
\caption{\textbf{Qualitative comparison on DTU dataset from 3 input images.} 
} \label{fig:dtu_qualitative} 
\vspace{-15pt}
\end{figure}

\begin{table*}[t]
\centering
\resizebox{\textwidth}{!}{%
\begin{tabular}{l|cccccccccccccccc}
\toprule
\textbf{Methods} & \textbf{scan24} & \textbf{scan37} & \textbf{scan40} & \textbf{scan55} & \textbf{scan63} & \textbf{scan65} & \textbf{scan69} & \textbf{scan83} & \textbf{scan97} & \textbf{scan105} & \textbf{scan106} & \textbf{scan110} & \textbf{scan114} & \textbf{scan118} & \textbf{scan122} & \textbf{$\downarrow$ Mean}\\
\midrule
MASt3R \cite{leroy2024grounding} & \cellcolor{tabsecond}2.49 & 20.48 & 12.02 & 5.39 & 8.86 & \cellcolor{tabthird}6.3 & 4.71 & 9.01 & 7.70 & 10.33 & 5.86 & \cellcolor{tabsecond}2.50 & 5.0 & \cellcolor{tabthird}4.56 & \cellcolor{tabthird}4.95 & 7.34\\
UFORecon \cite{na2024uforecon} & 35.55 & 57.15 & 38.81 & 35.47 & 50.02 & 44.94 & 34.70 & 53.56 & 41.67 & 47.93 & 40.22 & 39.34 & 39.75 & 33.34 & 49.04 & 42.77\\
CasMVSNet \cite{gu2020cascade} & 9.76 & 38.00 & 37.03 & 17.30 & 19.39 & 24.10 & 6.57 & 29.57 & 11.84 & 14.92 & 12.38 & 8.38 & \cellcolor{tabthird}3.93 & 9.87 & 13.44 & 17.10\\
Spurfies \cite{raj2024spurfies} & \cellcolor{tabthird}2.96 & 52.15 & 58.42 & 20.25 & 14.66 & 19.24 & 4.97 & 12.44 & \cellcolor{tabthird}7.50 & 10.64 & 7.36 & 6.26 & 38.61 & 25.84 & 44.19 & 21.70\\
InstSplatGOF \cite{fan2024instantsplat1} & 9.54 & 23.14 & 13.74 & 15.62 & 13.30 & 12.33 & 11.19 & 12.21 & 14.69 & 15.61 & 13.61 & 12.30 & 15.31 & 12.90 & 12.05 & 13.84\\
Sparsecraft \cite{younes2024sparsecraft} & 2.96 & \cellcolor{tabfirst}\textbf{10.51} & \cellcolor{tabfirst}\textbf{7.16} & \cellcolor{tabfirst}\textbf{2.92} & \cellcolor{tabthird}8.17 & 9.44 & \cellcolor{tabthird}3.83 & \cellcolor{tabthird}6.19 & 7.61 & \cellcolor{tabsecond}6.44 & \cellcolor{tabthird}4.75 & 6.43 & 5.36 & 7.65 & 8.14 & \cellcolor{tabthird}6.50\\
InstantSplat2DGS \cite{fan2024instantsplat2} & \cellcolor{tabfirst}\textbf{2.42} & \cellcolor{tabthird}21.74 & \cellcolor{tabthird}9.08 & \cellcolor{tabsecond}3.08 & \cellcolor{tabsecond}6.77 & \cellcolor{tabsecond}3.60 & \cellcolor{tabsecond}3.63 & \cellcolor{tabsecond}5.24 & \cellcolor{tabsecond}5.82 & \cellcolor{tabthird}7.12 & \cellcolor{tabsecond}4.48 & \cellcolor{tabthird}2.78 & \cellcolor{tabsecond}3.62 & \cellcolor{tabfirst}\textbf{3.11} & \cellcolor{tabsecond}3.43 & \cellcolor{tabsecond}5.73\\
\textbf{Ours} & \cellcolor{tabfirst}\textbf{2.42} & \cellcolor{tabsecond}12.313 & \cellcolor{tabsecond}7.9 & \cellcolor{tabthird}3.15 & \cellcolor{tabfirst}\textbf{6.05} & \cellcolor{tabfirst}\textbf{3.526} & \cellcolor{tabfirst}\textbf{3.48} & \cellcolor{tabfirst}\textbf{4.98} & \cellcolor{tabfirst}\textbf{5.29} & \cellcolor{tabfirst}\textbf{6.22} & \cellcolor{tabfirst}\textbf{4.39} & \cellcolor{tabfirst}\textbf{2.41} & \cellcolor{tabfirst}\textbf{3.46} & \cellcolor{tabsecond}3.32 & \cellcolor{tabfirst}\textbf{3.38} & \cellcolor{tabfirst}\textbf{4.82}\\
\bottomrule
\end{tabular}}
\vspace{-10pt}
\caption{\textbf{3-view reconstruction results on DTU dataset} The relative error (rel$\downarrow$) is reported for 15 scans. Best results are \textbf{bold}.}
\label{tab:rel}
\vspace{-10pt}
\end{table*}

\begin{table*}[t]
\centering
\resizebox{\textwidth}{!}{%
\begin{tabular}{l|cccccccccccccccc}
\toprule
\textbf{Methods} & \textbf{scan24} & \textbf{scan37} & \textbf{scan40} & \textbf{scan55} & \textbf{scan63} & \textbf{scan65} & \textbf{scan69} & \textbf{scan83} & \textbf{scan97} & \textbf{scan105} & \textbf{scan106} & \textbf{scan110} & \textbf{scan114} & \textbf{scan118} & \textbf{scan122} & \textbf{$\uparrow$ Mean}\\
\midrule
MASt3R \cite{leroy2024grounding} & \cellcolor{tabsecond}0.888 & \cellcolor{tabsecond}0.695 & \cellcolor{tabthird}0.844 & \cellcolor{tabthird}0.874 & \cellcolor{tabsecond}0.837 & \cellcolor{tabthird}0.839 & \cellcolor{tabsecond}0.851 & 0.754 & \cellcolor{tabsecond}0.824 & 0.745 & \cellcolor{tabthird}0.871 & \cellcolor{tabfirst}\textbf{0.875} & \cellcolor{tabthird}0.884 & \cellcolor{tabsecond}0.831 & \cellcolor{tabthird}0.846 & \cellcolor{tabthird}0.830\\
UFORecon \cite{na2024uforecon} & 0.362 & 0.278 & 0.506 & 0.469 & 0.277 & 0.372 & 0.383 & 0.361 & 0.362 & 0.338 & 0.424 & 0.366 & 0.332 & 0.402 & 0.339 & 0.371\\
CasMVSNet \cite{gu2020cascade} & 0.718 & 0.502 & 0.619 & 0.756 & 0.693 & 0.683 & 0.740 & 0.564 & 0.704 & 0.681 & 0.785 & 0.795 & 0.859 & 0.778 & 0.755 & 0.709\\
Spurfies \cite{raj2024spurfies} & 0.863 & 0.35 & 0.297 & 0.711 & 0.742 & 0.668 & 0.771 & 0.671 & 0.777 & 0.683 & 0.78 & 0.771 & 0.509 & 0.572 & 0.433 & 0.640\\
InstSplatGOF \cite{fan2024instantsplat1} & 0.774 & 0.581 & 0.689 & 0.696 & 0.746 & 0.750 & 0.710 & 0.658 & 0.682 & 0.653 & 0.695 & 0.724 & 0.710 & 0.663 & 0.650 & 0.692\\
Sparsecraft \cite{younes2024sparsecraft} & 0.841 & \cellcolor{tabthird}0.679 & 0.827 & 0.864 & 0.728 & 0.714 & 0.834 & \cellcolor{tabthird}0.756 & 0.746 & \cellcolor{tabthird}0.765 & 0.837 & 0.629 & 0.801 & 0.761 & 0.735 & 0.768\\
InstSplat2DGS \cite{fan2024instantsplat2} & \cellcolor{tabthird}0.882 & 0.616 & \cellcolor{tabsecond}0.86 & \cellcolor{tabfirst}\textbf{0.899} & \cellcolor{tabthird}0.834 & \cellcolor{tabsecond}0.865 & \cellcolor{tabthird}0.845 & \cellcolor{tabsecond}0.783 & \cellcolor{tabthird}0.810 & \cellcolor{tabsecond}0.773 & \cellcolor{tabsecond}0.882 & \cellcolor{tabthird}0.849 & \cellcolor{tabsecond}0.895 & \cellcolor{tabthird}0.827 & \cellcolor{tabsecond}0.855 & \cellcolor{tabsecond}0.832\\
\textbf{Ours} & \cellcolor{tabfirst}\textbf{0.904} & \cellcolor{tabfirst}\textbf{0.75} & \cellcolor{tabfirst}\textbf{0.877} & \cellcolor{tabsecond}0.896 & \cellcolor{tabfirst}\textbf{0.866} & \cellcolor{tabfirst}\textbf{0.887} & \cellcolor{tabfirst}\textbf{0.867} & \cellcolor{tabfirst}\textbf{0.799} & \cellcolor{tabfirst}\textbf{0.845} & \cellcolor{tabfirst}\textbf{0.789} & \cellcolor{tabfirst}\textbf{0.888} & \cellcolor{tabsecond}0.874 & \cellcolor{tabfirst}\textbf{0.902} & \cellcolor{tabfirst}\textbf{0.85} & \cellcolor{tabfirst}\textbf{0.865} & \cellcolor{tabfirst}\textbf{0.857}\\
\bottomrule
\end{tabular}}
\vspace{-10pt}
\caption{\textbf{3-view reconstruction results on DTU dataset} Normal consistency (NC$\uparrow$) is reported for 15 scans. Best results are \textbf{bold}.}
\label{tab:nc}
\vspace{-10pt}
\end{table*}

\begin{figure}[t] \centering
\centering 
\includegraphics[width=\linewidth]{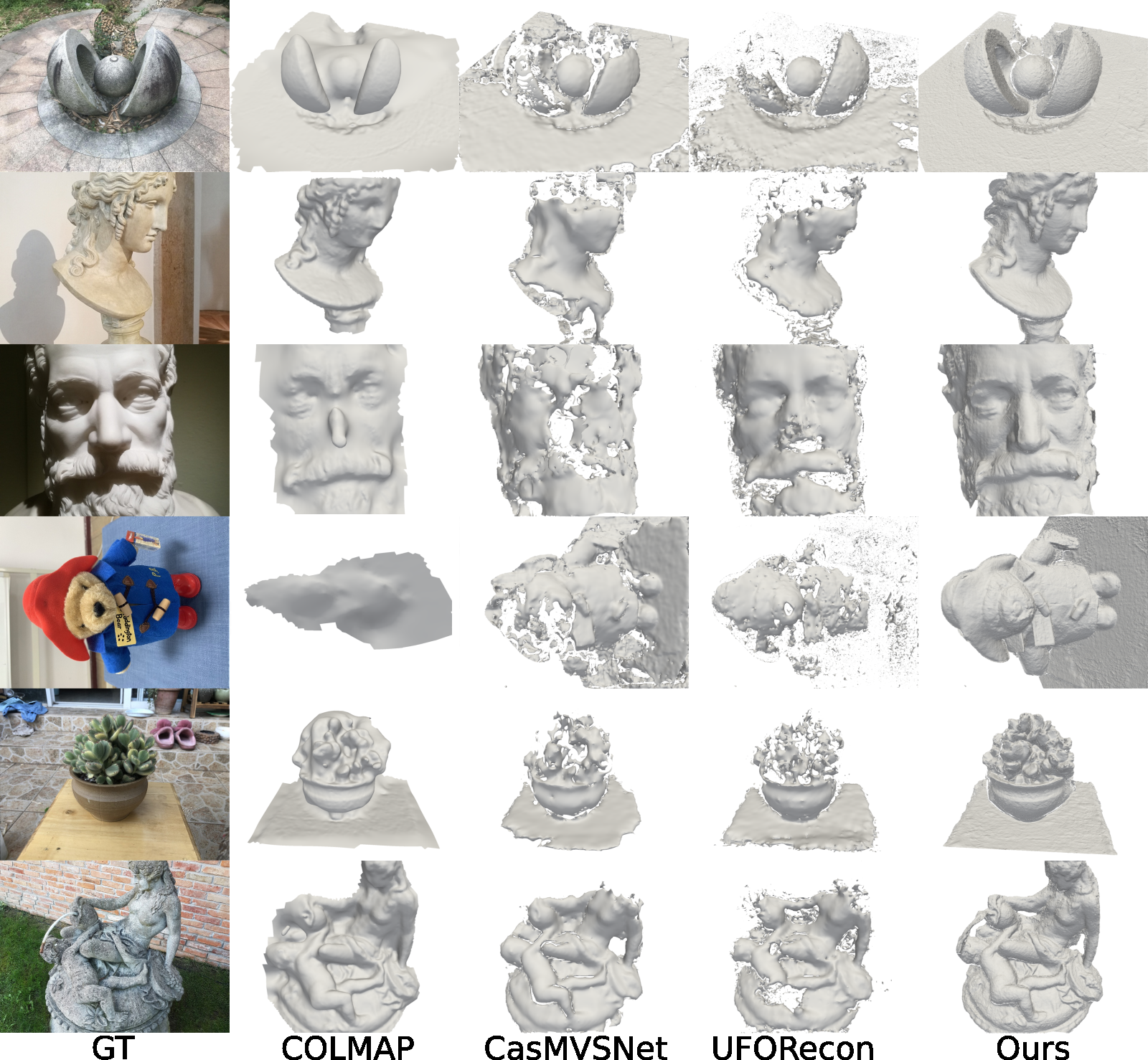} 
\vspace{-15pt}
\caption{\textbf{Qualitative comparison on BlendedMVS dataset from 3 input images.}} \label{fig:bmvs_qualitative} 
\vspace{-15pt}
\end{figure}


Tab.~\ref{tab:rel} and~\ref{tab:nc} present the quantitative results on the DTU dataset for Rel and NC metrics, respectively.
Fig \ref{fig:dtu_qualitative}, \ref{fig:bmvs_qualitative} and \ref{fig:mip_mvimg_qualitative} show qualitative reconstruction comparisons, with additional qualitative comparisons with SpaRP \cite{xu2024sparp} included in the supplementary material. Additionally, the supplementary material contains extended qualitative results, in the form of video comparisons demonstrating our reconstructions for several datasets. Our method achieves state-of-the-art performance, outperforming all competing approaches across the majority of scenes.

Several key observations can be made from these results. Pretrained generalizable methods (UfoRecon~\cite{na2024uforecon} and CasMVSNet~\cite{gu2020cascade}) generalize poorly when faced with noisy initial camera parameters from MASt3R~\cite{leroy2024grounding}, resulting in significantly worse performance. Despite being designed for sparse reconstruction, SparseCraft~\cite{younes2024sparsecraft} struggles with the noisy camera poses and point clouds from MASt3R~\cite{leroy2024grounding}. Spurfies~\cite{raj2024spurfies}, which employs an SDF prior trained on ShapeNet, performs particularly poorly with noisy camera poses, resulting in worse mesh extractions. InstantSplat \cite{fan2024instantsplat1, fan2024instantsplat2} variants (InstSplatGOF and InstSplat2DGS) perform better than other competing approaches, but still produce overly smooth meshes lacking fine details.

Our method significantly outperforms these approaches due to several key advantages, including the variance regularization loss (Sec.~\ref{ref:var_reg_section}) promoting sharper and more detailed reconstructions Fig \ref{fig:abl}, The correspondence loss (Sec.~\ref{ref:corr_section}) and joint optimization framework effectively refine the initial noisy camera poses from MASt3R~\cite{leroy2024grounding}. Our method's ability to preserve fine geometric details is also evidenced by the superior Normal Consistency scores.

\begin{table*}[t]
\centering
\resizebox{\textwidth}{!}{%
\begin{tabular}{l|cccccccccccccccc}
\toprule
\textbf{Methods} & \textbf{scan24} & \textbf{scan37} & \textbf{scan40} & \textbf{scan55} & \textbf{scan63} & \textbf{scan65} & \textbf{scan69} & \textbf{scan83} & \textbf{scan97} & \textbf{scan105} & \textbf{scan106} & \textbf{scan110} & \textbf{scan114} & \textbf{scan118} & \textbf{scan122} & \textbf{$\downarrow$ Mean}\\
\midrule
MASt3R \cite{leroy2024grounding} & 0.0202 & \cellcolor{tabthird}0.0689 & 0.0094 & 0.0203 & \cellcolor{tabthird}0.0171 & 0.0392 & \cellcolor{tabthird}0.0291 & 0.0294 & 0.0271 & 0.0326 & 0.0167 & \cellcolor{tabsecond}0.0210 & 0.0289 & 0.0192 & 0.0242 & 0.0269\\
InstSplatGOF \cite{fan2024instantsplat1} & \cellcolor{tabsecond}0.0180 & \cellcolor{tabsecond}0.0536 & \cellcolor{tabsecond}0.0089 & \cellcolor{tabsecond}0.0136 & \cellcolor{tabfirst}\textbf{0.0049} & \cellcolor{tabfirst}\textbf{0.0273} & \cellcolor{tabsecond}0.0263 & \cellcolor{tabfirst}\textbf{0.0239} & \cellcolor{tabsecond}0.0180 & \cellcolor{tabthird}0.0323 & \cellcolor{tabsecond}0.0128 & \cellcolor{tabthird}0.0238 & \cellcolor{tabsecond}0.0249 & \cellcolor{tabthird}0.0175 & \cellcolor{tabsecond}0.0149 & \cellcolor{tabsecond}0.0214\\
InstSplat2DGS \cite{fan2024instantsplat2} & \cellcolor{tabthird}0.0192 & 0.0895 & \cellcolor{tabthird}0.0090 & \cellcolor{tabthird}0.0154 & \cellcolor{tabsecond}0.0123 & \cellcolor{tabthird}0.0338 & 0.0294 & \cellcolor{tabsecond}0.0272 & \cellcolor{tabthird}0.0242 & \cellcolor{tabsecond}0.0322 & \cellcolor{tabthird}0.0166 & \cellcolor{tabsecond}0.0210 & \cellcolor{tabthird}0.0274 & \cellcolor{tabsecond}0.0163 & \cellcolor{tabthird}0.0175 & \cellcolor{tabthird}0.0261\\
\textbf{Ours} & \cellcolor{tabfirst}\textbf{0.0063} & \cellcolor{tabfirst}\textbf{0.0450} & \cellcolor{tabfirst}\textbf{0.0069} & \cellcolor{tabfirst}\textbf{0.0124} & 0.0175 & \cellcolor{tabsecond}0.0294 & \cellcolor{tabfirst}\textbf{0.0198} & \cellcolor{tabthird}0.0283 & \cellcolor{tabfirst}\textbf{0.0146} & \cellcolor{tabfirst}\textbf{0.0273} & \cellcolor{tabfirst}\textbf{0.0082} & \cellcolor{tabfirst}\textbf{0.0179} & \cellcolor{tabfirst}\textbf{0.0242} & \cellcolor{tabfirst}\textbf{0.0112} & \cellcolor{tabfirst}\textbf{0.0101} & \cellcolor{tabfirst}\textbf{0.0186}\\
\bottomrule
\end{tabular}}
\vspace{-10pt}
\caption{\textbf{Results on DTU dataset} Results include ATE$\downarrow$ metrics for 15 scans. 
Best results are \textbf{bold}.
}
\label{tab:ate}
\vspace{-10pt}
\end{table*}

Table~\ref{tab:ate} presents camera pose estimation results measured by ATE. Our approach achieves significant improvements in camera pose estimation, with a 13.1\% reduction in error compared to the best-performing baseline (InstSplatGOF \cite{fan2024instantsplat1}). This demonstrates the effectiveness of our correspondence loss in refining the initial noisy camera poses from MASt3R~\cite{leroy2024grounding}.

\subsection{Novel View Synthesis Evaluation} \label{subsec:nvs}

We evaluate our method's capability for novel view synthesis on Tanks and Temples~\cite{knapitsch2017tanks}, MVImgNet~\cite{yu2023mvimgnet}, and MipNeRF360~\cite{barron2022mip} datasets. 
Quantitative and qualitative result comparisons on all datasets are included in the supplementary material.

\begin{table}
\centering
\scriptsize
\begin{tabular}{l|ccc}
\toprule
\textbf{Methods} & \textbf{rel $\downarrow$} & \textbf{NC $\uparrow$}  & \textbf{ATE $\downarrow$}\\
\hline
2DGS & 7.93 & 0.785 & 0.0327\\
BA-2DGS & 7.79 & 0.788 & 0.0284\\
BA-2DGS+$\mathcal{L}_{\text{corr}}$ & \cellcolor{tabthird}7.41 & \cellcolor{tabthird}0.804 & \cellcolor{tabthird}0.0256\\
BA-2DGS+$\mathcal{L}_{\text{corr}}$+Norm. init. & \cellcolor{tabsecond}7.03 & \cellcolor{tabsecond}0.829 & \cellcolor{tabfirst}\textbf{0.0237}\\
BA-2DGS+$\mathcal{L}_{\text{corr}}$+Norm. init.+$\mathcal{L}_{\text{var}}$ & \cellcolor{tabfirst}\textbf{6.79} & \cellcolor{tabfirst}\textbf{0.834} & \cellcolor{tabsecond}0.0238\\
\bottomrule
\end{tabular}
\caption{\textbf{Ablation on DTU dataset, adding one component at a time.} Results include rel and NC metrics. Best results are \textbf{bold}.
}
\label{tab:abl_loss}
\end{table}

\subsection{Ablation Studies} 
\label{subsec:ablation}
We conduct ablation studies to analyze the contribution of each component in our method, including an experiment to study the impact of varying the number of input views.

\begin{figure} 
\centering 
\includegraphics[width=\linewidth]{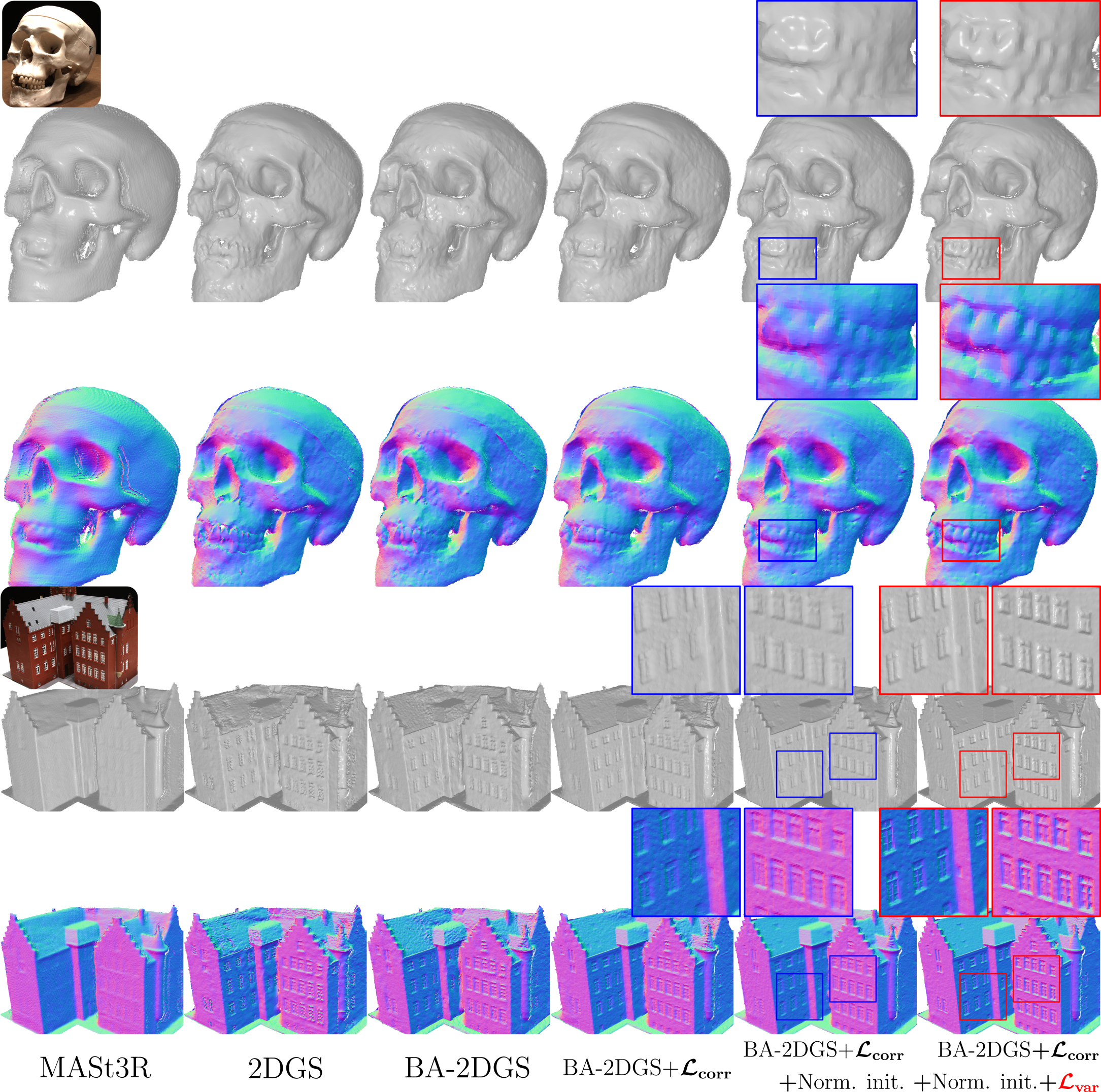}
\vspace{-15pt}
\caption{\textbf{Qualitative ablation of the component of our method, showing also the effect of variance loss.} Left blue box: w/o $\mathcal{L}_\text{var}$. Right red box: w $\mathcal{L}_\text{var}$. (3 input images). } \label{fig:abl} 
\vspace{-17pt}
\end{figure}

\vspace{-10pt}
\paragraph{Component Ablation}
These experiments were performed using 3 input unposed views on 5 testing scenes from the DTU dataset using two sets of views, following the protocols established in~\cite{long2022sparseneus,na2024uforecon}.
Tab~\ref{tab:abl_loss} shows the impact of incorporating different components into our pipeline. Fig \ref{fig:abl} shows a qualitative ablation comparison. Camera optimization using bundle adjustment improves all metrics. Incorporating the correspondence loss further enhances this alignment, resulting in improved reconstructions. Splat orientation initialization with pointmap normals improves results further. Notice that our variance loss contributes to both the final metrics and the qualitative aspect of our results. The reconstructions with variance loss exhibit significantly sharper details and better preservation of fine structures, validating the quantitative improvements observed in our metrics.

\begin{figure} \centering \includegraphics[width=1.15\linewidth]{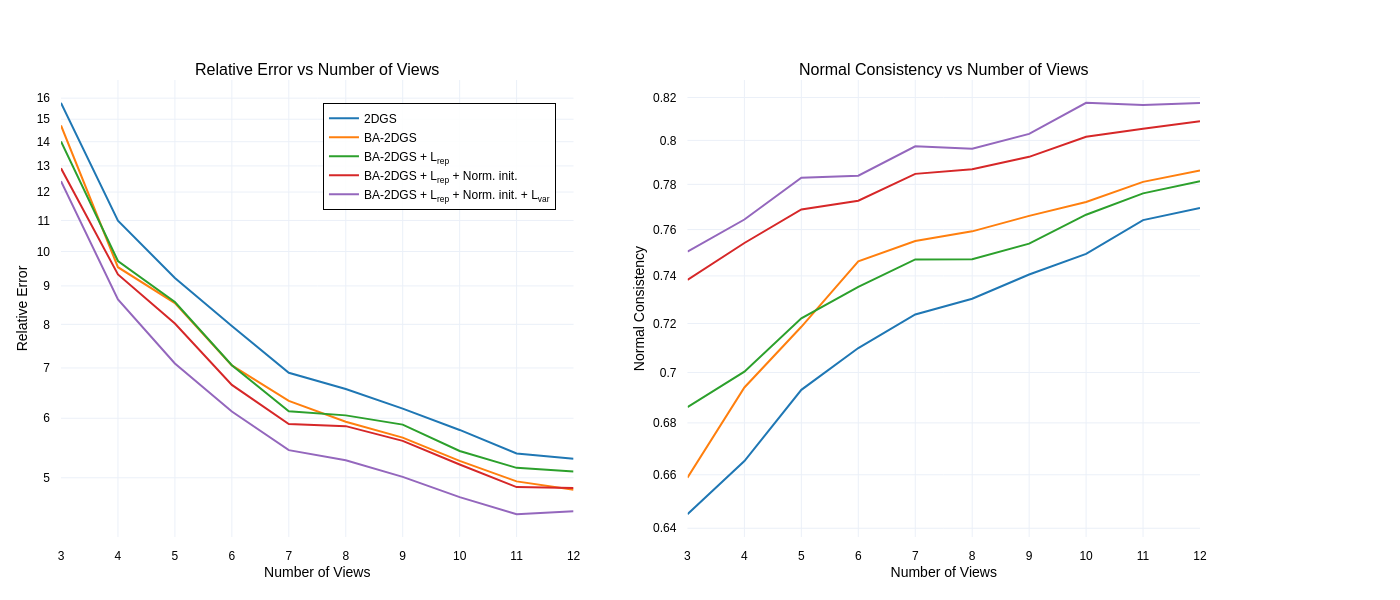} \caption{\textbf{Effect of increasing the number of input views on different metrics.}} \label{fig:abl_no_views} \end{figure}

\vspace{-10pt}
\paragraph{Number of Views Ablation}
We progressively increase the number of views and perform pose-free reconstruction with our method using scan37 from DTU. We evaluate by computing the mean metrics across both test view sets. As shown in Figure~\ref{fig:abl_no_views}, our method benefits favorably from increased views. The benefits of all components of our method are also maintained, if not increased. This includes our variance loss. 


%% file: sec/conc.tex
\vspace{-5pt}
\section{Running time} 
\vspace{-5pt}
In our reconstruction experiments, we observed that the mean training time on DTU \cite{aanaes2016large} for 1000 iterations (similar to InstantSplat-XL \cite{fan2024instantsplat1, fan2024instantsplat2}) was roughly 146s on a single Nvidia A6000 GPU. For a fair comparison, we also trained InstantSplat-XL \cite{fan2024instantsplat1, fan2024instantsplat2} on the same GPU and observed that the mean training time was around 128s. Hence, our training time is in the same range as InstantSplat \cite{fan2024instantsplat1, fan2024instantsplat2} while offering significantly higher reconstruction and novel-view quality.

\section{Limitations and Future Work} 
\label{sec:limitations}

\begin{figure}[h] \centering
\centering 
\includegraphics[width=\linewidth]{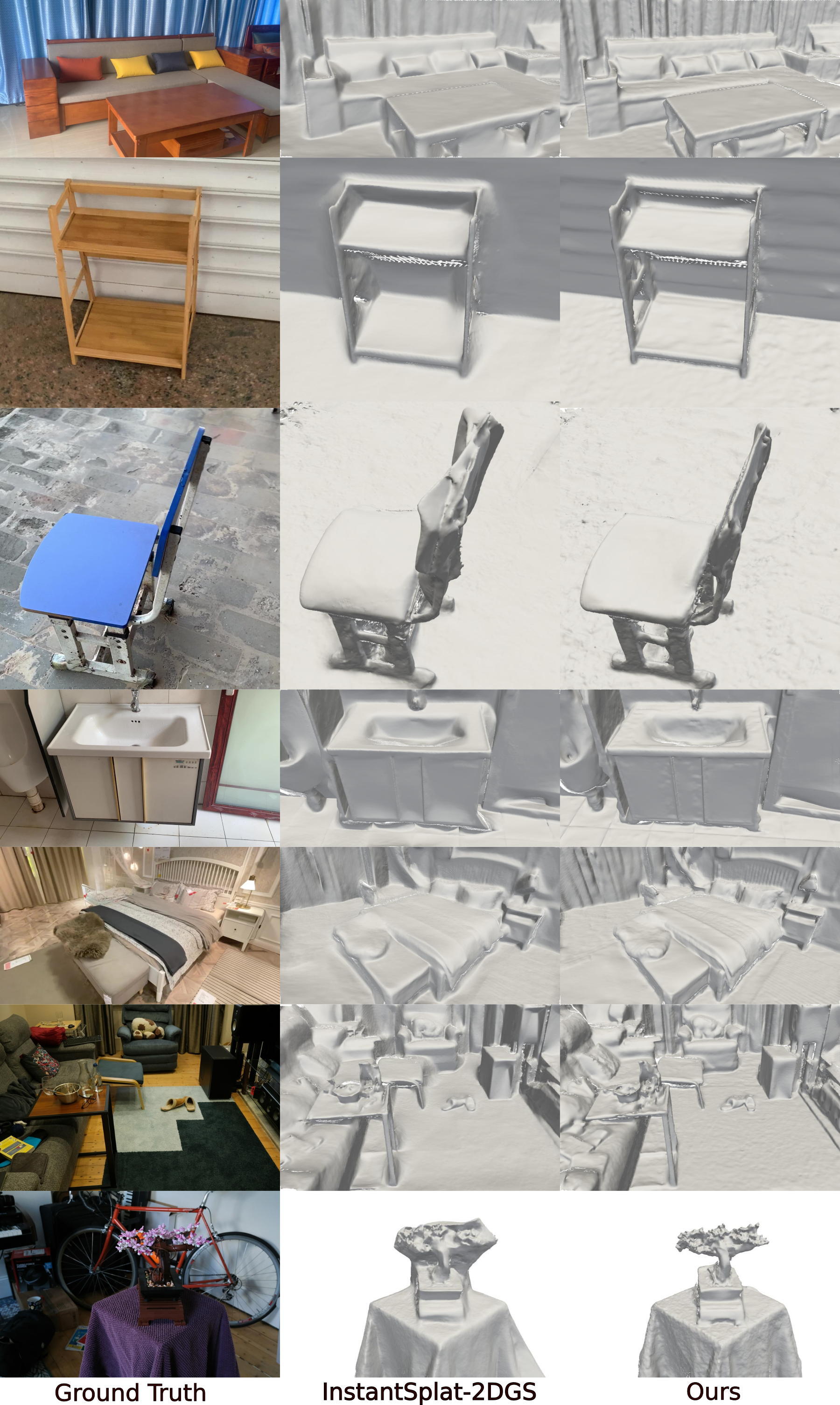}
\vspace{-20pt}
\caption{\textbf{Qualitative comparison on MVImgNet and MipNeRF360 datasets from 6 and 12 input images, respectively.}} \label{fig:mip_mvimg_qualitative} 
\end{figure}

Despite the promising results, several limitations remain, which present opportunities for future research. As foundation models keep on improving, better initializations could be available in the future. Furthermore, while our approach handles a variety of scenes across different datasets, performance may be limited for highly complex geometry, very thin structures, or semi-transparent objects (observed for the glass bottle in Fig \ref{fig:mip_mvimg_qualitative} room scene in MipNeRF360~\cite{barron2022mip}), which are common challenges for surface based methods especially in the sparse setting. Additionally, our method assumes a static scene under consistent lighting. Extending it to more dynamic scenarios is one of the challenges we are looking forward to exploring. 

\section{Conclusion}

We addressed the underexplored problem of shape reconstruction from sparse uncalibrated multi-view images. Our method leverages foundation models for initial camera and geometry estimation, followed by joint optimization of 2D Gaussian primitives and camera parameters. The introduction of a variance regularization loss significantly improves the detail and sharpness of reconstructions, while our correspondence matching based framework enhances camera pose accuracy. Extensive evaluations on multiple datasets demonstrate that our approach consistently outperforms state-of-the-art methods in reconstruction quality, novel view synthesis, and camera pose estimation. 


%% file: sec/supp.tex
\section{Evaluation of Novel View Synthesis and Camera pose estimation}
\begin{figure} \centering \includegraphics[width=\linewidth]{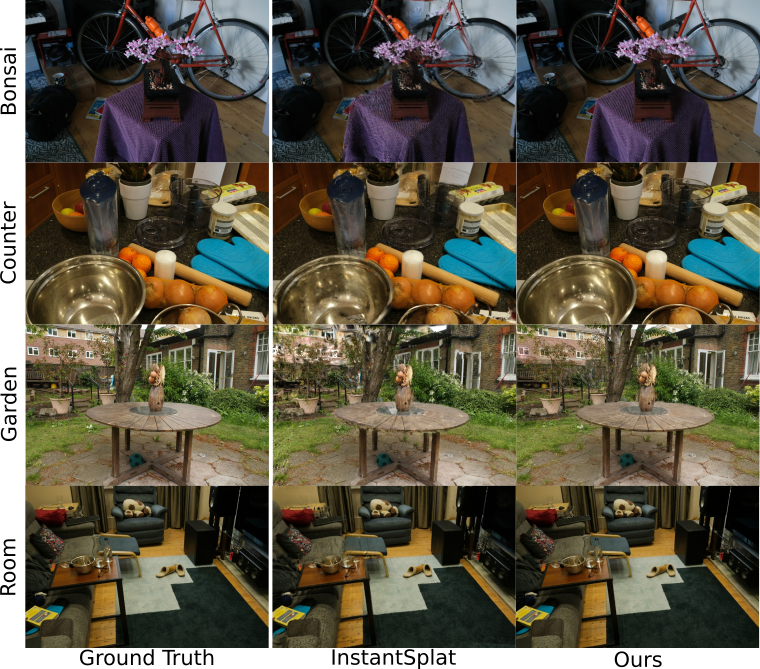} 
\vspace{-15pt}
\caption{\textbf{Qualitative comparison of novel view synthesis on MipNeRF360 dataset from 12 input images.}
} \label{fig:nvs_qualitative_mip} 
\end{figure}

\begin{figure} 
\centering 
\includegraphics[width=\linewidth]{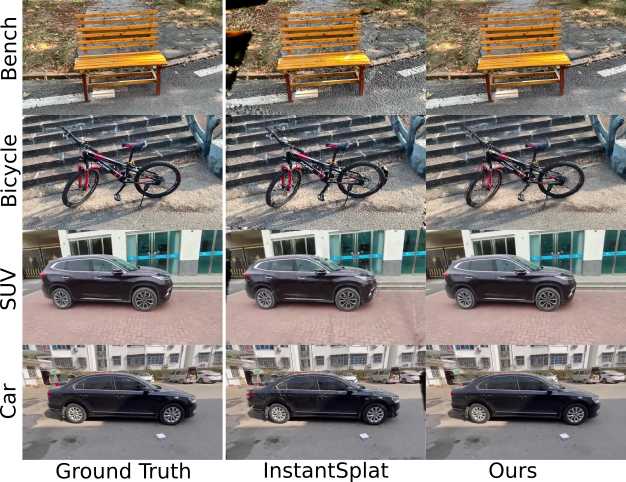} 
\caption{\textbf{Qualitative comparison of novel view synthesis on MVImgNet dataset from 3 input images.}} \label{fig:nv_mvimg} 
\end{figure}

\begin{figure} 
\centering 
\includegraphics[width=\linewidth]{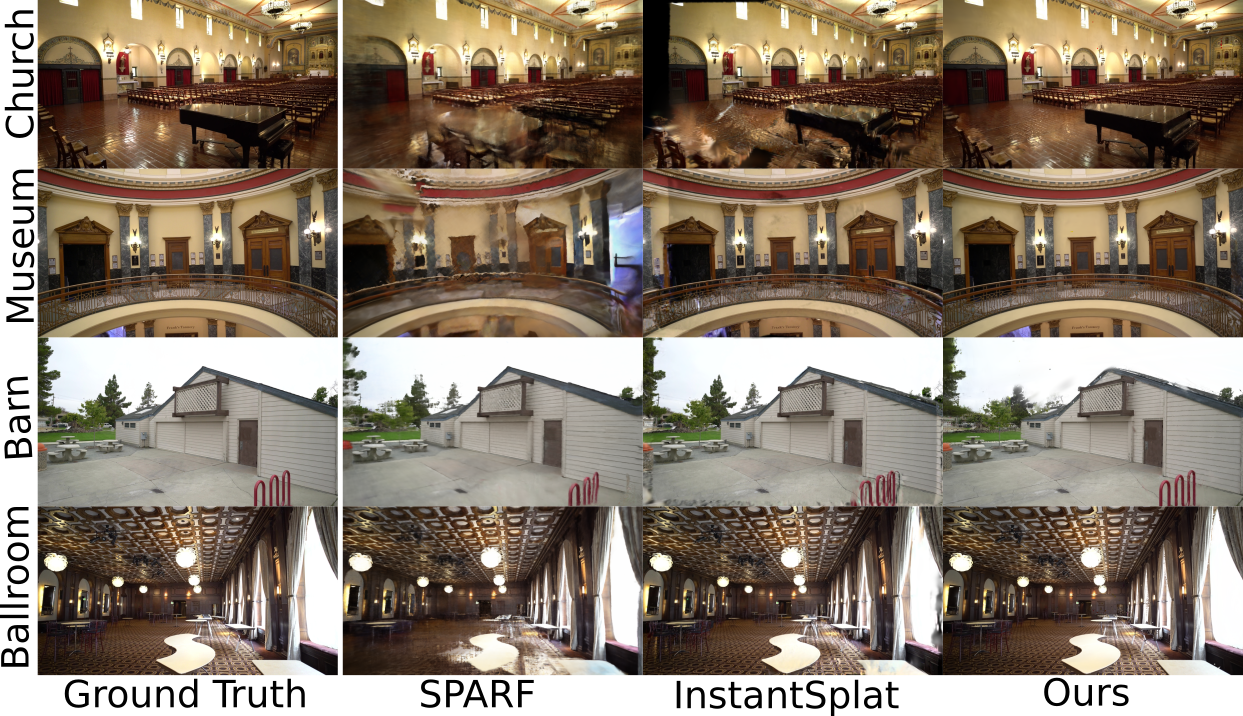} 
\caption{\textbf{Qualitative comparison of novel view synthesis on Tanks and Temples dataset from 3 input images.}} \label{fig:nv_tnt} 
\end{figure}

In this section, we provide qualitative and quantitative comparisons on the Tanks and Temples \cite{knapitsch2017tanks}, MipNeRF360~\cite{barron2022mip} datasets and MVImgNet \cite{yu2023mvimgnet} datasets for both novel view synthesis and camera pose estimation metrics. Tab \ref{tab:mvimg_mipnerf_nvs} presents quantitative results for these experiments on MipNeRF360~\cite{barron2022mip} and MVImgNet~\cite{yu2023mvimgnet} with qualitative results presented in Fig \ref{fig:nvs_qualitative_mip} and Fig \ref{fig:nv_mvimg}. For Tanks and Temples \cite{knapitsch2017tanks}, quantitative and qualitative results are reported in Tab \ref{tab:tnt_nvs} and Fig \ref{fig:nv_tnt} respectively. We observe that NoPe-NeRF~\cite{bian2023nope} and NeRF-mm \cite{wang2021nerf} suffer markedly in their novel view performance and camera pose estimation metrics. Being implicit, volumetric rendering methods, they also suffer from slow training and inference times. CF-3DGS~\cite{fu2023colmap} also encounters artifacts when rendering from novel viewpoints, stemming from its complex optimization pipeline and erroneous pose estimations. InstantSplat \cite{fan2024instantsplat1,fan2024instantsplat2} variants provide good performance, but still lag behind our method in most metrics, particularly in the challenging 3-view setting. For the Tanks and Temples comparison in Tab \ref{tab:tnt_nvs}, Fig \ref{fig:nv_tnt}, we also outperform SPARF \cite{truong2023sparf} by a sizeable margin on all metrics, while requiring order of magnitudes less training and inference time, since it takes around 10 hours to train on a single scene and needs more than a minute to render a single image during inference, owing to its volumetric rendering framework. Our method significantly outperforms all baselines on various datasets in terms of SSIM, LPIPS (novel view synthesis metrics) and ATE (camera pose estimation metric), demonstrating its robustness to complex scenes with challenging lighting conditions.

\begin{table}[ht]
\centering
\resizebox{1.0\linewidth}{!}{%
\begin{tabular}{l|ccc|ccc|ccc|cccc}
\hline
\multirow{2}{*}{\textbf{Method}} & \multicolumn{3}{c|}{\textbf{SSIM (MVImgNet)}} & \multicolumn{3}{c|}{\textbf{LPIPS (MVImgNet)}} & \multicolumn{3}{c|}{\textbf{ATE (MVImgNet) $\downarrow$}} & \multicolumn{4}{c}{\textbf{MipNeRF360 (12 Training Views)}} \\
\cline{2-14}
& \textbf{3-view} & \textbf{6-view} & \textbf{12-view} & \textbf{3-view} & \textbf{6-view} & \textbf{12-view} & \textbf{3-view} & \textbf{6-view} & \textbf{12-view} & \textbf{SSIM} & \textbf{PSNR} & \textbf{LPIPS} & \textbf{ATE $\downarrow$} \\
\hline
NoPe-NeRF \cite{bian2023nope} & 0.4326 & 0.4329 & 0.4686 & 0.6168 & 0.6614 & 0.6257 & 0.2780 & \cellcolor{tabthird}0.1740 & 0.1493 & 0.3580 & 16.16 & 0.6867 & 0.2374 \\
CF-3DGS \cite{fu2023colmap} & 0.3414 & 0.3544 & 0.3655 & 0.4520 & 0.4326 & 0.4492 & \cellcolor{tabthird}0.1593 & 0.1981 & 0.1243 & 0.2443 & 13.17 & 0.6098 & 0.2263 \\
NeRF-mm \cite{wang2021nerf} & 0.3752 & 0.3685 & 0.3718 & 0.6421 & 0.6252 & 0.6020 & 0.2721 & 0.2376 & 0.1529 & 0.2003 & 11.53 & 0.7238 & 0.2401 \\
Instantsplat-S \cite{fan2024instantsplat1,fan2024instantsplat2} & \cellcolor{tabthird}0.5489 & \cellcolor{tabthird}0.6835 & \cellcolor{tabthird}0.7050 & \cellcolor{tabthird}0.3941 & \cellcolor{tabthird}0.2980 & \cellcolor{tabthird}0.3033 & \cellcolor{tabfirst}\textbf{0.0184} & \cellcolor{tabsecond}0.0259 & \cellcolor{tabsecond}0.0165 & \cellcolor{tabsecond}0.4647 & \cellcolor{tabsecond}17.68 & \cellcolor{tabthird}0.5027 & \cellcolor{tabsecond}0.2161 \\
Instantsplat-XL \cite{fan2024instantsplat1,fan2024instantsplat2} & \cellcolor{tabsecond}0.5628 & \cellcolor{tabsecond}0.6933 & \cellcolor{tabsecond}0.7321 & \cellcolor{tabsecond}0.3688 & \cellcolor{tabsecond}0.2611 & \cellcolor{tabsecond}0.2421 & \cellcolor{tabfirst}\textbf{0.0184} & \cellcolor{tabsecond}0.0259 & \cellcolor{tabfirst}\textbf{0.0164} & \cellcolor{tabthird}0.4398 & \cellcolor{tabthird}17.23 & \cellcolor{tabsecond}0.4486 & \cellcolor{tabthird}0.2162 \\
\hline
\textbf{Ours} & \cellcolor{tabfirst}\textbf{0.8313} & \cellcolor{tabfirst}\textbf{0.8801} & \cellcolor{tabfirst}\textbf{0.9008} & \cellcolor{tabfirst}\textbf{0.2215} & \cellcolor{tabfirst}\textbf{0.1658} & \cellcolor{tabfirst}\textbf{0.1410} & \cellcolor{tabsecond}0.0273 & \cellcolor{tabfirst}\textbf{0.0244} & \cellcolor{tabthird}0.0172 & \cellcolor{tabfirst}\textbf{0.8168} & \cellcolor{tabfirst}\textbf{26.21} & \cellcolor{tabfirst}\textbf{0.2199} & \cellcolor{tabfirst}\textbf{0.2067} \\
\hline
\end{tabular}%
}
\vspace{-10pt}
\caption{NVS Performance Comparison of Different Methods on MVImgNet and MipNeRF360}
\label{tab:mvimg_mipnerf_nvs}
\vspace{-10pt}
\end{table}

\begin{table}
\centering
\resizebox{1.0\linewidth}{!}{%
\begin{tabular}{l|ccc|ccc|ccc}
\hline
\textbf{Method} & \multicolumn{3}{c|}{\textbf{SSIM}$\uparrow$} & \multicolumn{3}{c|}{\textbf{LPIPS}$\downarrow$} & \multicolumn{3}{c}{\textbf{ATE}$\downarrow$}\\ 
& \textbf{3-view} & \textbf{6-view} & \textbf{12-view} & \textbf{3-view} & \textbf{6-view} & \textbf{12-view} & \textbf{3-view} & \textbf{6-view} & \textbf{12-view}\\ \hline
COLMAP + 3DGS \cite{kerbl20233d} & 0.3755 & 0.5917 & 0.7163 & 0.5130 & 0.3433 & 0.2505 & - & - & -\\ 
COLMAP + FSGS \cite{zhu2023fsgs} & 0.5701 & 0.7752 & 0.8479 & 0.3465 & 0.1927 & 0.1477 & - & - & -\\ 
NoPe-NeRF \cite{bian2023nope} & 0.4570 & 0.5067 & 0.6096 & 0.6168 & 0.5780 & 0.5067 & 0.2828 & 0.1431 & 0.1029\\ 
CF-3DGS \cite{fu2023colmap} & 0.4066 & 0.4690 & 0.5077 & 0.4520 & 0.4219 & 0.4189 & 0.1937 & 0.1572 & 0.1031\\ 
NeRF-mm \cite{wang2021nerf} & 0.4019 & 0.4308 & 0.4677 & 0.6421 & 0.6252 & 0.6020 & 0.2721 & 0.2329 & 0.1529\\ 
SPARF \cite{truong2023sparf} & 0.5751 & 0.6731 & 0.5708 & 0.4021 & 0.3275 & 0.4310 & 0.0568 & 0.0554 & 0.0385\\ 
Instantsplat-S \cite{fan2024instantsplat2} & \cellcolor{tabsecond}0.7624 & \cellcolor{tabthird}0.8300 & \cellcolor{tabthird}0.8413 & \cellcolor{tabthird}0.1844 & \cellcolor{tabthird}0.1579 & \cellcolor{tabthird}0.1654 & \cellcolor{tabthird}0.0191 & \cellcolor{tabsecond}0.0172 & \cellcolor{tabthird}0.0110\\ 
Instantsplat-XL \cite{fan2024instantsplat2} & \cellcolor{tabthird}0.7615 & \cellcolor{tabsecond}0.8453 & \cellcolor{tabsecond}0.8785 & \cellcolor{tabsecond}0.1634 & \cellcolor{tabfirst}\textbf{0.1173} & \cellcolor{tabsecond}0.1068 & \cellcolor{tabsecond}0.0189 & \cellcolor{tabfirst}\textbf{0.0164} & \cellcolor{tabsecond}0.0101\\ 
\textbf{Ours} & \cellcolor{tabfirst}\textbf{0.8752} & \cellcolor{tabfirst}\textbf{0.9020} & \cellcolor{tabfirst}\textbf{0.9180} & \cellcolor{tabfirst}\textbf{0.1623} & \cellcolor{tabsecond}0.1283 & \cellcolor{tabfirst}\textbf{0.1050} & \cellcolor{tabfirst}\textbf{0.0150} & \cellcolor{tabthird}0.0174 & \cellcolor{tabfirst}\textbf{0.0078}\\ \hline
\end{tabular}%
}
\caption{Performance comparison of different methods across SSIM, LPIPS, and ATE metrics for 3-view, 6-view, and 12-view settings on the Tanks and Temples dataset.}
\label{tab:tnt_nvs}
\end{table}

\section{Additional qualitative comparison on 3D reconstruction}
\begin{figure} 
\centering 
\includegraphics[width=\linewidth]{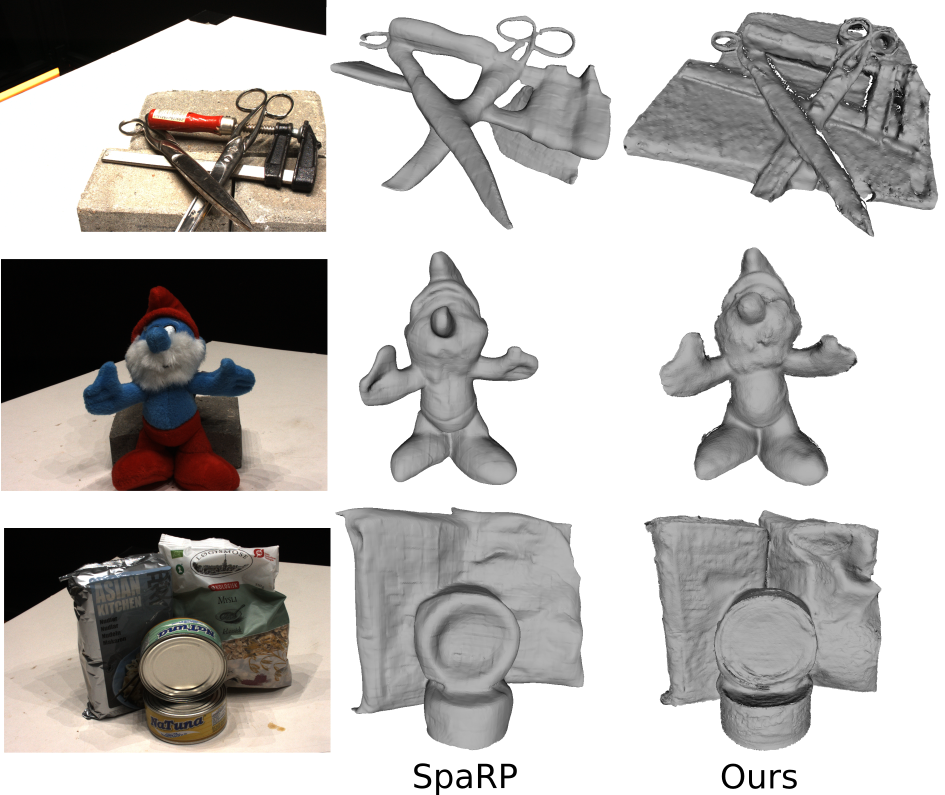} 
\caption{\textbf{Qualitative comparison with SpaRP \cite{xu2024sparp} on DTU from 3 input images.}} \label{fig:sparp_dtu_qualitative} 
\end{figure}

We also provide a qualitative comparison in Fig. \ref{fig:sparp_dtu_qualitative} to SpaRP \cite{xu2024sparp} on the DTU \cite{aanaes2016large} dataset, a recent method that leverages 2D diffusion models for efficient 3D reconstruction and pose estimation from unposed sparse-view images. For comparison using 3 input images, our method achieves mesh reconstructions with greater fidelity to the input images, as seen in the comparisons. We also provide video results depicting our reconstructions and novel view results on the DTU \cite{aanaes2016large} and BlendedMVS \cite{yao2020blendedmvs} datasets and their comparison to other methods.

\section{Color variance plot} 
\begin{figure}[t]
 \vspace{-9pt} 
 \centering
  \includegraphics[width=0.5\textwidth]{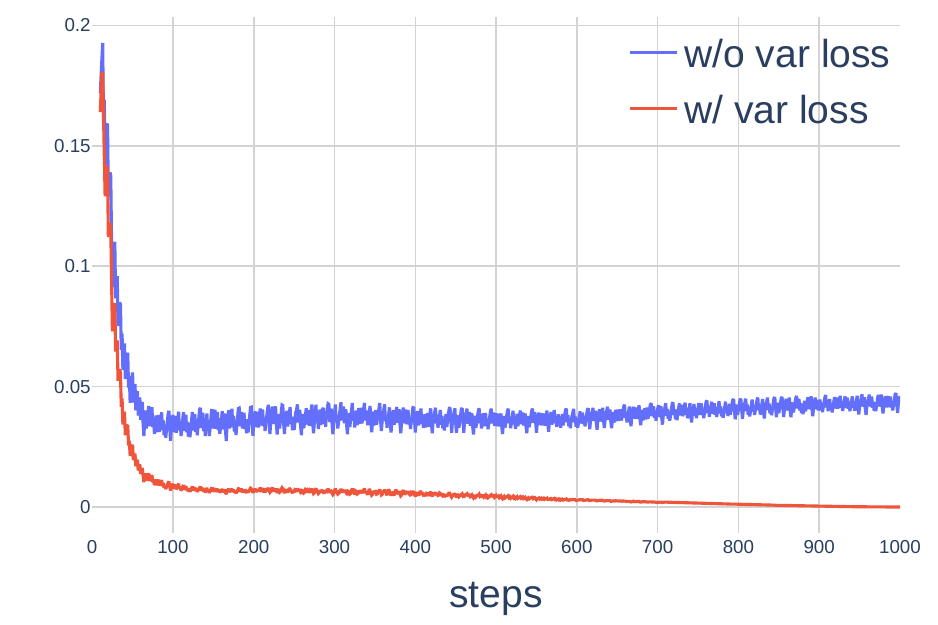}
  \vspace{-15pt}
  \caption{ 
\textbf{Color variance.} The variance loss keeps color variance lower, encouraging stable, robust convergence.
}
\label{fig:var_plot}
\end{figure}

We plot the average color variance over optimization iterations (Fig. \ref{fig:var_plot}) for models w/ and w/o variance loss. Models with the loss activated effectively maintain lower color variance consistently, which aligns with our goal of encouraging stable, low-uncertainty renderings. This supports the effectiveness of the proposed loss in guiding convergence toward robust geometry.

\section{Alternative priors} 
\begin{figure}[h]
 \vspace{-12pt} 
 \centering
  \includegraphics[width=0.5\textwidth]{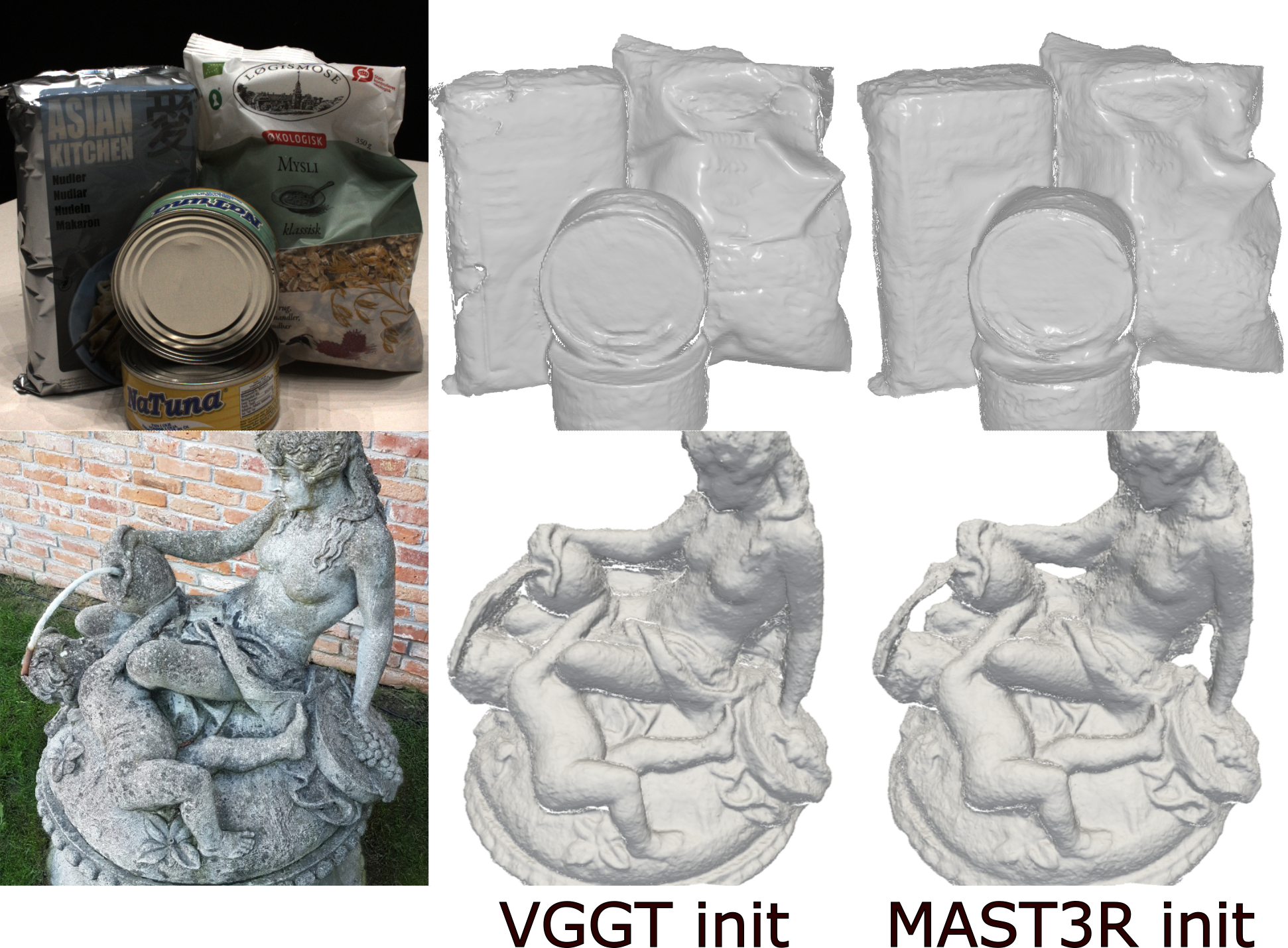}
  \vspace{-15pt}
  \caption{ 
\textbf{Alternative prior.} VGGT initialization shows successful results (3 input images), demonstrating our framework’s modularity.
}
\label{fig:prior_plot}
\end{figure}

This example (Fig. \ref{fig:prior_plot}) demonstrates that initializing our framework with VGGT \cite{wang2025vggt}, a recent state-of-the-art feed-forward method that avoids the global optimization step of MASt3R \cite{leroy2024grounding} also produces successful results. This highlights the modularity of our approach and its compatibility with different geometric priors.

\section{Variance loss motivation} Our goal is to hedge against epistemic uncertainty in geometry estimation inherent to the unposed surface reconstruction problem. Under sparse-view supervision, the rendering objective admits many geometries that fit the training images but generalizes poorly (see Sec.3 in \cite{zhang2020nerf++}). This issue is exacerbated when camera poses are optimized during training, introducing additional noise into the supervision. In Gaussian Splatting, scene geometry is encoded through splat parameters defining the 3D density. Among the many plausible geometries, we seek to bias the model toward those that remain predictive even under small perturbations, \ie robust solutions less sensitive to noisy supervision signals. To formalize this, we minimize the worst-case deviation in rendered color under perturbations to the geometric density field (Eq. \ref{eqn:var_loss1}), yielding a variance regularization loss (Eq. \ref{eqn:var_loss3}) that penalizes color variance along rays. From a learning-theoretic perspective, this can be interpreted as seeking flat minima \cite{hochreiter1997flat} in the space of densities, an idea supported by arguments from both statistical and deep learning viewpoints, and shown to be effective across a range of machine learning applications \cite{chaudhari2019entropy, mackay1992practical, hochreiter1997flat, foret2020sharpness}. Empirically, this leads to more stable and consistent reconstructions from sparse views (Fig. \ref{fig:abl}, Tab. \ref{tab:abl_loss}).

\section{Prior failures} 
The example below illustrates a typical scenario where MASt3R's \cite{leroy2024grounding} feed-forward geometry prediction struggles: reconstruction from only 6 images without known camera poses. Challenging regions such as texture-less surfaces, highly reflective materials (e.g., glass doors and shiny faucets), and thin structures often lead to noisy or incomplete results. In contrast, our method recovers plausible geometry in these cases thanks to robust test-time optimization, which refines both the pose and the reconstructed shape despite imperfect initializations.
\begin{figure}[h]
\centering
\vspace{-10pt}
\includegraphics[width=1.0\linewidth]{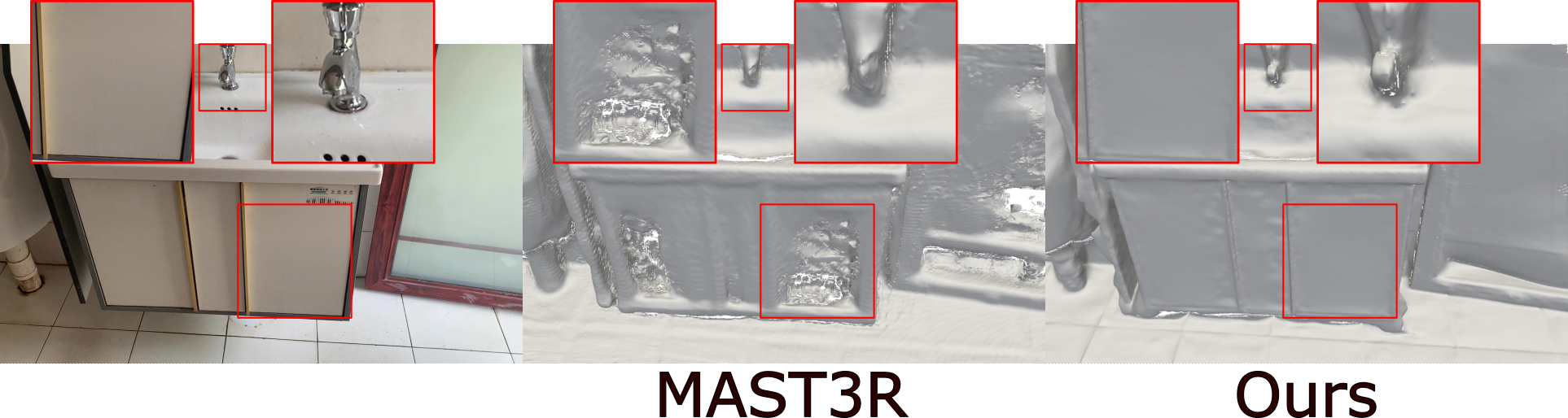}
\vspace{-15pt}
\caption{\textbf{Robustness to MASt3R failure (3 input images).} Our method recovers geometry where MASt3R struggles.}
\end{figure}